\documentclass[10pt,twocolumn,letterpaper]{article}

\usepackage[pagenumbers]{cvpr} 

\usepackage[dvipsnames]{xcolor}

\definecolor{cvprblue}{rgb}{0.21,0.49,0.74}
\usepackage[pagebackref,breaklinks,colorlinks,citecolor=cvprblue]{hyperref}

\usepackage{tabularx}
\usepackage{multirow}
\usepackage{dblfloatfix}

\newcommand{\best}[1]{{\color{blue} \textbf{#1}}}
\DeclareMathOperator*{\argmin}{arg\,min}
\newcolumntype{Y}{>{\centering\arraybackslash}X}

\def\paperID{64} 
\def\confName{3DV\xspace}
\def\confYear{2024\xspace}

\title{SUCRe: Leveraging Scene Structure for Underwater Color Restoration}

\author{
Clémentin Boittiaux\textsuperscript{1,2,3}\quad Ricard Marxer\textsuperscript{3}\quad Claire Dune\textsuperscript{2}\quad Aurélien Arnaubec\textsuperscript{1}\\
Maxime Ferrera\textsuperscript{1}\quad Vincent Hugel\textsuperscript{2}\\
\textsuperscript{1} Ifremer, Centre Méditerranée\quad \textsuperscript{2} Université de Toulon, COSMER\\
\textsuperscript{3} Université de Toulon, Aix Marseille Univ, CNRS, LIS
}

\begin{document}
\maketitle
\begin{abstract}
Underwater images are altered by the physical characteristics of the medium through which light rays pass before reaching the optical sensor. Scattering and wavelength-dependent absorption significantly modify the captured colors depending on the distance of observed elements to the image plane. In this paper, we aim to recover an image of the scene as if the water had no effect on light propagation. We introduce SUCRe, a novel method that exploits the scene's 3D structure for underwater color restoration. By following points in multiple images and tracking their intensities at different distances to the sensor, we constrain the optimization of the parameters in an underwater image formation model and retrieve unattenuated pixel intensities. We conduct extensive quantitative and qualitative analyses of our approach in a variety of scenarios ranging from natural light to deep-sea environments using three underwater datasets acquired from real-world scenarios and one synthetic dataset. We also compare the performance of the proposed approach with that of a wide range of existing state-of-the-art methods. The results demonstrate a consistent benefit of exploiting multiple views across a spectrum of objective metrics. Our code is publicly available at \href{https://github.com/clementinboittiaux/sucre}{github.com/clementinboittiaux/sucre}.
\end{abstract}

\section{Introduction}

Images captured under the water are significantly different from those taken above the surface. Water has a large effect on light transport, inducing various changes of appearance in the scene. Particles in suspension propagate light in multiple directions, inducing what is known as scattering. In addition, strong wavelength-dependent absorption greatly modifies the color of light that reaches the sensor. More importantly, these effects are highly dependent on the water conditions and the distance between elements of the scene and the sensor~\cite{solonenko2015jerlov, akkaynak2017attenuation, akkaynak2018revised, mcglamery1980model, jaffe1990modeling}. This source of visual variability presents a challenge for interpreting and exploiting underwater images and has been long sought to be reduced.

\begin{figure}[t]
    \centering
    \includegraphics[width=\linewidth]{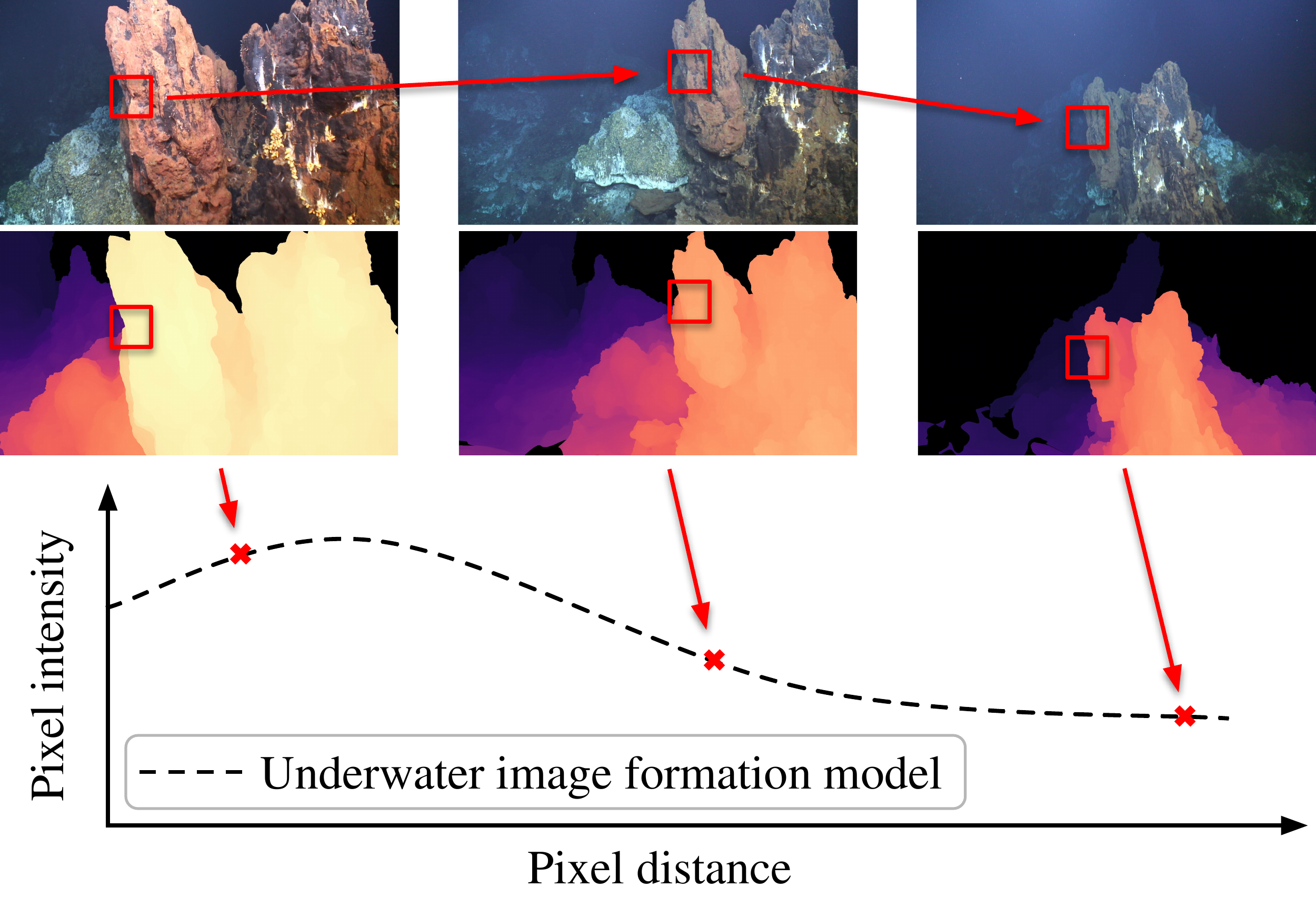}
    \caption{\textbf{Multi-view tracking.} We track pixels in multiple images to retrieve their intensities at different distances. We then estimate simultaneously their corrected color and the parameters of an underwater image formation model.}
    \label{fig:method}
\end{figure}

Underwater color restoration aims to remove the effects of the water medium on light transport, a strategy that has already been shown to significantly improve the usability of underwater images in many tasks~\cite{ancuti2017dehazing, berman2021haze, li2016restoration}. We can identify two main types of approaches employed to address this problem: statistical methods and physics-based models. Statistical methods rely on intrinsic properties of the images to improve their visual aspect. Some of these methods rely on fusing color and contrast enhanced versions of the original underwater image~\cite{ancuti2012fusion, ancuti2017dehazing}, while others leverage the expressiveness of neural networks by training models using pairs of underwater images and their corresponding references. Such pairs of images can be generated synthetically~\cite{zwilgmeyer2021varos}, obtained via domain distribution modeling such as GANs~\cite{li2018watergan} or selected using human preference manual annotations~\cite{li2020benchmark}. In contrast, physics-based approaches have focused on explicitly modeling the formation of images under water~\cite{jaffe1990modeling, mcglamery1980model, akkaynak2017attenuation, akkaynak2018revised, nakath2021restoration}, invert the model and recover the scene appearance as if it were captured without the effects of the described phenomena~\cite{akkaynak2019seathru, berman2021haze, berman2017haze, nakath2021restoration, li2018watergan}. For instance, \textit{Sea-thru}~\cite{akkaynak2019seathru} restores colors from a single image and its corresponding distance map by estimating the parameters related to absorption and scattering. However, estimating these parameters from a single image is an ill-posed problem. To cope with this, \textit{Sea-thru} has to rely on additional assumptions, like a distance-wise adaptation of the dark channel prior~\cite{he2010haze}, to allow independent estimation of the parameters.

We hereby propose to exploit multiple views from the same scene to further constrain the aforementioned ill-posed problem. We rely on Structure-from-Motion (SfM) not only to estimate distance maps, but also to track observed intensities of elements in the scene at different distances based on estimated camera poses. In recent years, SfM has become a common intermediate step of many popular airborne computer vision algorithms~\cite{mildenhall2020nerf,sarlin2019hloc}. In underwater scenarios, SfM has been proven highly feasible with an appropriate camera setup~\cite{campos2015eiffeltower, johnsonroberson2010sfm, ferrera2019aqualoc}. Pizarro \etal have demonstrated the effectiveness of underwater SfM, even in scenarios with low overlapping images and challenging 3D structures~\cite{pizarro2009reconstruction}. As such, SfM has also become increasingly popular for use in underwater computer vision algorithms~\cite{nakath2021restoration, sethuraman2022waternerf, skinner20173dcorrection, akkaynak2019seathru, bryson2015color, levy2023seathrunerf}. While it requires additional information about the scene, the proposed approach is well suited for the creation of large-scale datasets with reference images from real-world scenarios, which can be used to train underwater image enhancement neural networks.

Our contributions are the following:\\
\textbf{1)} We introduce a novel multi-view method that simultaneously estimates the parameters of an underwater image formation model alongside the restored image by tracking points in multiple images to retrieve their intensities at different distances to the scene (see \cref{fig:method}).\\
\textbf{2)} We validate experimentally the developed approach on synthetic and real-world datasets in both natural light and deep-sea scenarios. We perform extensive objective quantitative evaluation on two datasets containing reference ground truth data: synthesized underwater images~\cite{zwilgmeyer2021varos} and color charts captured under water~\cite{akkaynak2019seathru}. The applicability and ecological validity of our method is confirmed with qualitative analysis on images from two deep dive surveys: Eiffel Tower~\cite{matabos2015eiffel} and a submarine wreck. Our results demonstrate that leveraging multiple views enables restoring colors that are significantly attenuated in the underwater images due to the distance to the scene. Furthermore, it consistently improves the accuracy of color rendering of elements. Given its improved performance, our approach can be used to produce new reference images for training single-view underwater color restoration methods. To fully appreciate the effectiveness of our underwater color restoration results, we encourage readers to watch the accompanying supplementary videos, which provide compelling visual comparisons.

\section{Related work}\label{sec:related}

A large body of work tackles restoration of single still images without using any additional source of information~\cite{berman2021haze, berman2017haze, hou2020dcp, zhou2021restoration, chiang2012dehazing, schechner2005polarization, li2016restoration}. While single image restoration is the most general and challenging formulation of the problem, work on the multi-view setting is of great benefit to produce approximate references for evaluation or target pairs for supervised learning, \eg, in neural network-based approaches \cite{li2018watergan, li2020benchmark, pipara2021color}.

In this context, previous works have focused on restoring the colors of underwater 3D models. For instance, after generating an underwater 3D model using SfM, Bryson \etal leverage multi-view observations of texture patches to estimate the parameters of an underwater image formation model and restore the color of the model's texture map~\cite{bryson2012colour, bryson2015color}. On the other hand, Nakath \etal~\cite{nakath2021restoration} propose employing a differentiable rendering framework, where they use the parameters of a previously fitted underwater image formation model to render a textured 3D mesh from multiple views. The error between the acquired and the rendered images is then backpropagated directly into the texture map. In contrast, the proposed approach focuses on restoring any full-resolution image registered in the SfM by directly minimizing a multi-view objective within the image itself, eliminating limitations such as the resolution of texture maps and enabling color restoration on small datasets consisting of only a few overlapping images.

Our work relies on two main prior developments that are presented here as background.

\begin{table}
    \begin{center}
    \begin{tabularx}{\linewidth}{cXl}
    \toprule
    Variable & Description & Type \\
    \midrule
    $\pmb{I}$ & underwater images & $\mathbb{R}^{N \times H \times W \times C}$ \\
    $\pmb{J}$ & restored images & $\mathbb{R}^{N \times H \times W \times C}$ \\
    $\pmb{z}$ & distance maps of images & $\mathbb{R}^{N \times H \times W}$ \\
    $B$ & veiling light  & $\mathbb{R}^{C}$ \\
    $i$ & image index & $\left[1..N\right]$ \\
    $c$ & color channel index & $\left[1..C\right]$ \\
    $p$ & pixel index & $\left[1..H \times W\right]$ \\
    \bottomrule
    \end{tabularx}
    \end{center}
    \caption{\textbf{Underwater image formation model variables used across the paper.} We use subscripts to index specific images, color channels and pixels, \eg, $\pmb{I}_{i,c,p}$ is the intensity of pixel $p$ of channel $c$ of image $i$. In some equations the image index $i$ may be omitted for readability. Bold symbols are used for variables encoding spatial information such as images or distances maps. $N$ is the number of images (\ie, views), $H$ and $W$ are the height and width of images and $C$ the number of channels $C = \left|\{R, G, B\}\right|$.}
    \label{tab:variables}
\end{table}

\begin{figure*}[t]
    \centering
    \includegraphics[width=\linewidth]{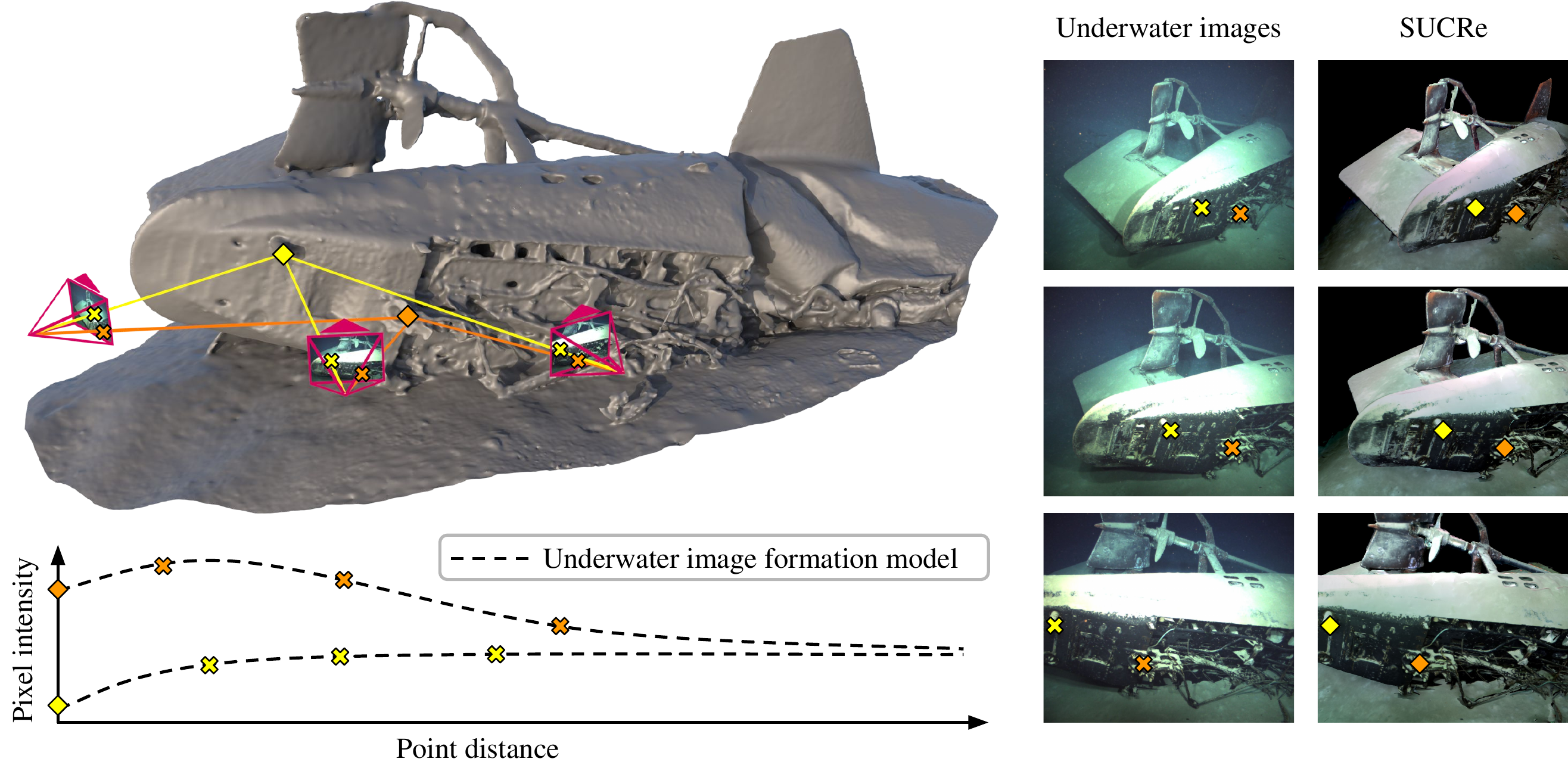}
    \caption{\textbf{SUCRe pipeline.} We use camera poses, intrinsics and depth maps resulting from a SfM to pair geometrically pixels between different views. We project pixels from one view to another, enabling us to pair points in low contrast areas. We then simultaneously estimate an UIFM parameters along with the restored image. This figure illustrates our method on a real-world deep-sea dive at a submarine wreck.}
    \label{fig:pipeline}
\end{figure*}

\paragraph{Underwater Image Formation Model (UIFM):} A key component of physics-based color restoration methods is their UIFM that describes how the colors of the observed scene are affected by the water medium~\cite{jaffe1990modeling, mcglamery1980model, akkaynak2017attenuation, akkaynak2018revised, nakath2021restoration}. While there are many phenomena that impact the quality of underwater images, two of them are predominant: \textit{i)} light collides with floating particles that then act as sources of light, inducing scattering~\cite{akkaynak2018revised}; \textit{ii)} light is attenuated through the water column~\cite{akkaynak2017attenuation}. Because both phenomena strongly depend on the wavelength~\cite{akkaynak2017attenuation, akkaynak2018revised}, UIFM parameters are often wavelength-dependent. In most cases, color restoration is performed for each color channel independently~\cite{schechner2005polarization, akkaynak2019seathru, nakath2021restoration, berman2017haze, berman2021haze, song2022survey}.

Many underwater color restoration methods~\cite{berman2021haze, berman2017haze, schechner2005polarization, chiang2012dehazing} rely on the UIFM introduced in~\cite{schechner2005polarization} to model backscatter and color attenuation in natural light conditions:
\begin{equation}
  \pmb{I}_{c,p} = \pmb{J}_{c,p} e^{-\alpha_c \pmb{z}_p} + B_c (1 - e^{-\alpha_c \pmb{z}_p}),
  \label{eq:haze_model}
\end{equation}
where $\alpha \in \mathbb{R}^C$ is the wavelength-dependent coefficient weighting the distance dependency of color attenuation and backscatter. The other variables are described in \cref{tab:variables}.

Akkaynak \etal~\cite{akkaynak2017attenuation, akkaynak2018revised} further revised this model to account for differences between backscatter and attenuation coefficients:
\begin{equation}
  \pmb{I}_{c,p} = \pmb{J}_{c,p} e^{-\beta_c \pmb{z}_p} + B_c (1 - e^{-\gamma_c \pmb{z}_p}),
  \label{eq:seathru}
\end{equation}
where $\beta \in \mathbb{R}^C$ is the color attenuation coefficient and $\gamma \in \mathbb{R}^C$ is the backscatter coefficient.

\paragraph{\textit{Sea-thru}} is a state-of-the-art underwater image color restoration method that relies on images in a raw file format and their corresponding distance maps~\cite{akkaynak2019seathru}. It focuses on inverting the UIFM described by \cref{eq:seathru}. Given that the distance maps are generated using SfM, there is potential to leverage the scene's 3D information in order to constrain the estimation of the parameters in \cref{eq:seathru}.

\section{Method}\label{sec:methods}

In \textit{Sea-thru}, with the help of the distance information, the problem has $\lvert \pmb{I}_c \rvert$ equations and $\lvert \pmb{I}_c \rvert + \lvert k \rvert$ unknowns, with $k = \{\beta_c, B_c, \gamma_c\}$ the set of UIFM parameters. Given there are more unknowns than observations, the problem is underdetermined and requires additional assumptions to constrain the optimization. For example, an extreme trivial solution can be found with $\pmb{J}_c \to \pmb{I}_c$, $\beta_c \to 0$ and $B_c \to 0$. To tackle this, \textit{Sea-thru} relies on a distance-based alternative to the dark channel prior~\cite{he2010haze} to retrieve $B_c$ and $\gamma_c$, and an illuminant map estimation~\cite{ebner2013color} to retrieve $\beta_c$. $\pmb{J}_c$ is then retrieved from \cref{eq:seathru} using these parameters:
\begin{equation}
  \pmb{J}_{c,p} = \left(\pmb{I}_{c,p} - B_c (1 - e^{-\gamma_c \pmb{z}_p})\right) e^{\beta_c \pmb{z}_p}.
  \label{eq:j}
\end{equation}

This paper introduces a novel approach, named SUCRe, that overcomes the limitations of single-view underwater image color restoration methods by leveraging multiple observations of the scene, thus eliminating the need for additional assumptions. Our method takes as input undistorted underwater images together with their corresponding camera poses, intrinsics and depth maps. This information is retrieved using an off-the-shelf SfM pipeline~\cite{sarlin2019hloc}. By pairing pixels in multiple images, we are able to follow the intensity evolution of points at different distances and estimate the parameters of the UIFM and pixel intensities at a hypothetical zero meter distance, implying a lack of disturbance by the water medium (see \cref{fig:pipeline}).

\paragraph{SfM pipeline:}
As a preliminary step to our approach, we outline the procedure to obtain the inputs for SUCRe using SfM. We adopt the pipeline proposed by Sarlin \etal for visual localization, which is designed to be robust to significant changes in the environment~\cite{sarlin2019hloc}. NetVLAD~\cite{arandjelovic2016netvlad} pairs similar images between which pixels are matched using SuperPoint~\cite{detone2018superpoint} features and SuperGlue~\cite{sarlin2020superglue} matcher. The bundle adjustment is performed with COLMAP SfM~\cite{schoenberger2016sfm} to recover the camera poses and intrinsic parameters. We then undistort the images using the estimated intrinsic parameters and build a 3D mesh of the scene. Finally, depth maps are obtained by ray-casting the images onto the 3D mesh. This results in depth maps less noisy than those obtained by multi-view stereo methods.

\begin{figure}[t]
    \centering
    \includegraphics[width=\linewidth]{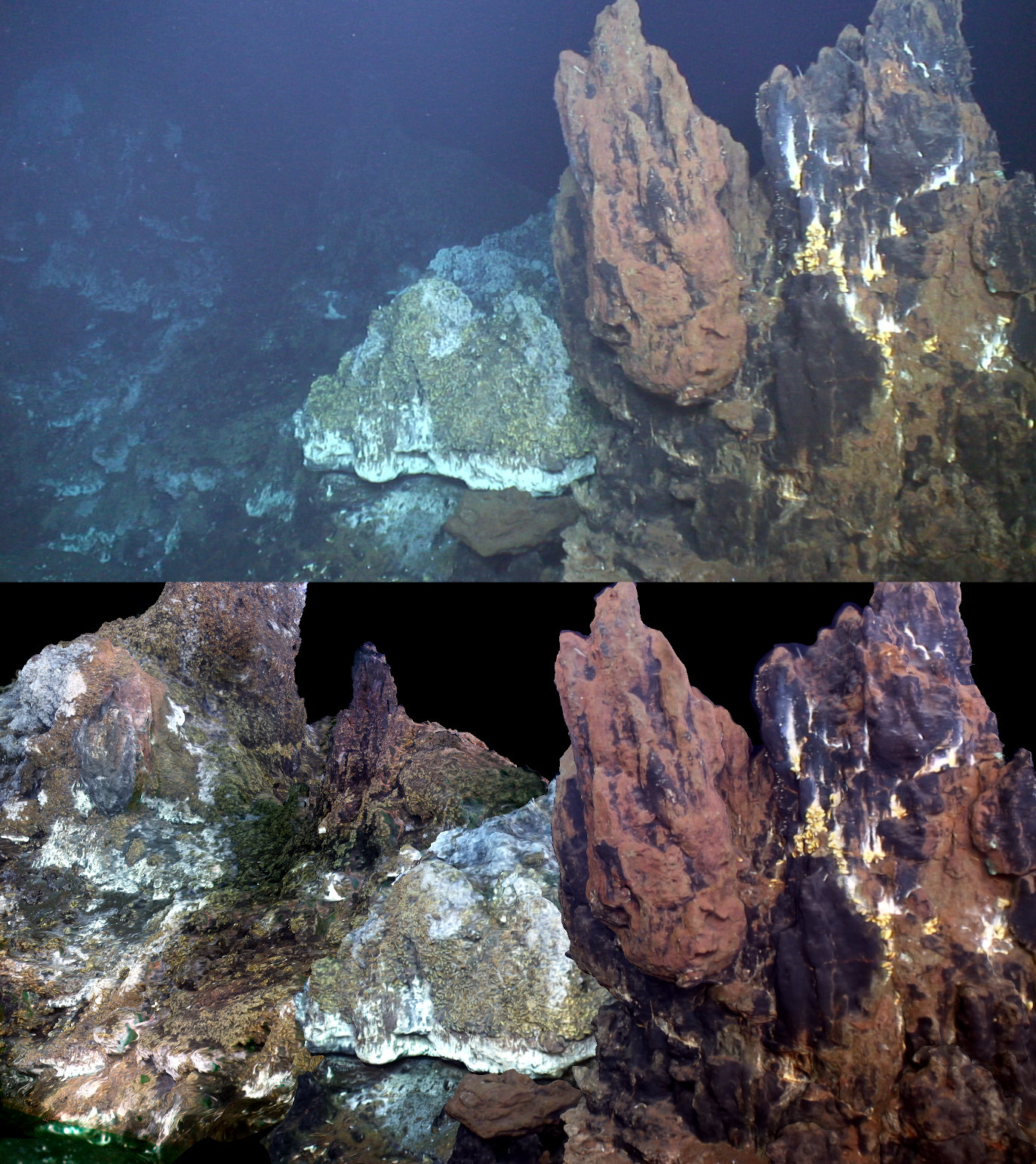}
    \caption{\textbf{Applying SUCRe on a deep-sea image} from the Eiffel Tower dataset~\cite{matabos2015eiffel,boittiaux2023eiffeltower} captured by a ROV equipped with an artificial lighting system. The figure depicts the recovery of colors in low contrast areas (top left of the image). Pixels without depth information are left blank.}
    \label{fig:wow}
\end{figure}

\paragraph{Dense multi-view pixel pairing:}
The first step of our approach is to pair pixels in a dense manner between different views. This is accomplished by projecting pixel coordinates from one view to another using the depth maps as well as the poses of the cameras and their intrinsics parameters. Let $x_1$ be the homogeneous coordinates of a pixel in image view $i_1$ with depth $d_1 \in \mathbb{R}$ and homogeneous pose matrix $^{w}\pmb{T}_{i_1} \in SE(3)$. The pose matrix $^{w}\pmb{T}_{i_1}$ represents the rigid transformation from frame $i_1$ to frame $w$ such as: $^w\lambda = \! ^{w}\pmb{T}_{i_1} \odot\, ^{i_1}\lambda$, where $^{i_1}\lambda \in \mathbb{R}^{3}$ and $^w\lambda \in \mathbb{R}^{3}$. Let $\pmb{K} \in \mathbb{R}^{3 \times 3}$ be the intrinsic calibration matrix of images $i_1$ and $i_2$. The projection of $x_1$ in image view $i_2$ with pose $^{w}\pmb{T}_{i_2}$ can be obtained in homogeneous coordinates by:
\begin{equation}
  x_2 = \pmb{K} \, ^{i_2}\pmb{T}_{w} \, \odot \left( ^{w}\pmb{T}_{i_1} \odot \pmb{K}^{-1} d_1 \, x_1 \right).
  \label{eq:match}
\end{equation}
We then back-project $x_2$ in $i_1$ view using $i_2$ depth map:
\begin{equation}
  x_1^{\prime} = \pmb{K} \, ^{i_1}\pmb{T}_{w} \odot \left( \, ^{w}\pmb{T}_{i_2} \odot \pmb{K}^{-1} d_2 \, x_{2} \right),
  \label{eq:match_reverse}
\end{equation}
where $d_2$ is the depth of $x_2$ in image view $i_2$. The pixels in both images ($x_1$, $x_2$) are only paired if $x_1^{\prime}$ and $x_1$ land on the same pixel coordinate, \ie, pixels are matched to each other in both directions, from $i_1$ to $i_2$ and from $i_2$ to $i_1$. This ensures that each pixel has only one match in both images. It also filters out points occluded by the structure. This geometry-based approach allows us to robustly pair pixels in scenarios where feature matching algorithms fail, \eg, in low contrast areas where most image signal has been attenuated like in the top left corner of \cref{fig:wow}.

\paragraph{Optimization:}

With multiple observations of the same $\pmb{J}_{c,p}$, our problem becomes well-posed. In SUCRe, we formulate the model described by \cref{eq:seathru} in a multi-view setting:
\begin{equation}
  \pmb{I}_{i,c,p} = \pmb{J}_{c,p} e^{-\beta_c \pmb{z}_{i,p}} + B_c (1 - e^{-\gamma_c \pmb{z}_{i,p}}).
  \label{eq:sucre_uifm}
\end{equation}
Parameters of \cref{eq:sucre_uifm} are then estimated by fitting the model in a least squares manner:
\begin{multline}
    \argmin_{\pmb{J}_{c}, B_c, \beta_c, \gamma_c} \sum_i \sum_p \Vert \pmb{I}_{i,c,p} - \pmb{J}_{c,p} e^{-\beta_c \pmb{z}_{i,p}} \\
    - B_c (1 - e^{-\gamma_c \pmb{z}_{i,p}}) \Vert^2.
    \label{eq:sfmseathru_likelihood}
\end{multline}
In \Cref{ap:ols}, we offer insights that justify the selection of the least squares estimator for estimating the model's parameters.

Because some pixels in low contrast areas were matched with closer observations (see \cref{fig:pipeline}), we are able to retrieve their color despite insufficient information about them on the image being restored (see \cref{fig:wow}). Pixels with depth that have not been paired in other images contribute to the optimization as single observations.

\begin{table*}
    \begin{center}
    \small
    \begin{tabularx}{\linewidth}{p{4.5cm}ccccXcccc}
    \toprule
    \multirow{2}{*}[-2.5pt]{Method} & \multicolumn{4}{c}{Varos} & & \multicolumn{4}{c}{\textit{Sea-thru} D5} \\
    \cmidrule(lr){2-5} \cmidrule(lr){7-10}
    & PSNR $\uparrow$ & SSIM $\uparrow$ & UCIQE $\uparrow$ & UIQM $\uparrow$ & & $\bar{\psi}$ $\downarrow$ & $\bar{\psi}$ std $\downarrow$ & $\Delta E_{00}$ $\downarrow$ & $\Delta E_{00}$ std $\downarrow$ \\
    \midrule
    Underwater image & 10.71 & 0.39 & 0.60 & 1.40 & & 37.14 & 3.72 & 36.93 & 3.68 \\
    \cmidrule(lr){2-10}
    Fusion~\cite{ancuti2012fusion} & 10.25 & 0.35 & 0.51 & 2.10 & & 29.85 & 6.38 & 30.60 & 6.34 \\
    \textit{Sea-thru*}~\cite{akkaynak2019seathru} & 10.15 & 0.39 & 0.52 & 1.88 & & 27.55 & 3.68 & 30.64 & 5.46 \\
    Water-Net~\cite{li2020benchmark} & 11.20 & 0.38 & 0.54 & 1.96 & & 29.12 & 4.11 & 31.49 & 5.89 \\
    FUnIE-GAN~\cite{islam2020funie} & 11.02 & 0.35 & \best{0.62} & 2.51 & & 32.91 & 3.63 & 35.55 & 5.07 \\
    Haze-Lines~\cite{berman2021haze} & 9.64 & 0.36 & 0.57 & 2.00 & & 25.80 & 7.14 & 28.85 & 6.89 \\
    TACL~\cite{liu2022tacl} & 10.02 & 0.36 & 0.44 & \best{2.52} & & 29.28 & 4.27 & 30.50 & 4.93 \\
    \textbf{SUCRe (ours)} & \best{12.13} & \best{0.42} & 0.32 & 1.99 & & \best{21.45} & \best{2.63} & \best{22.56} & \best{2.84} \\
    \bottomrule
    \end{tabularx}
    \end{center}
    \caption{\textbf{Restoration evaluation.} Our approach shows significant improvements across all full-reference metrics in comparison to other methods. Additionally, its achieves high performance independently of the color chart position, as indicated by the lower standard deviation on \textit{Sea-thru} D5 errors.}
    \label{tab:results}
\end{table*}

\section{Experiments}\label{sec:experiments}

This work has been developed in the context of deep-sea applications. However, the image formation model described by \cref{eq:seathru}, which we extended in a multi-view setting in \cref{eq:sucre_uifm}, was conceived for natural light conditions. In this section, we demonstrate the effectiveness of our approach in both natural light and deep-sea environments. We show that while the employed model does not capture some specific characteristics of deep-sea imaging, it is sufficient to outperform existing solutions. For more details, \Cref{ap:fitting} provides empirical examples that illustrate how the natural light model fits deep-sea images. Taking these factors into consideration, the design and application of deep-sea image formation models requires further research and remains in the scope of future works.

\subsection{Datasets}

We evaluate our method on four distinct datasets. Two of these datasets, Varos~\cite{zwilgmeyer2021varos} and \textit{Sea-thru} D5~\cite{akkaynak2019seathru}, offer respectively reference images and color charts with known colors, enabling the calculation of quantitative metrics. The remaining two datasets consist of ecologically valid deep-sea data, showcasing the real-life practical applicability of our method for two sites of interest, \ie, the Eiffel Tower hydrothermal vent~\cite{matabos2015eiffel,boittiaux2023eiffeltower} and a submarine wreck. Real-world datasets were acquired with camera housings that account for air-glass-water refraction, allowing the use of SfM.



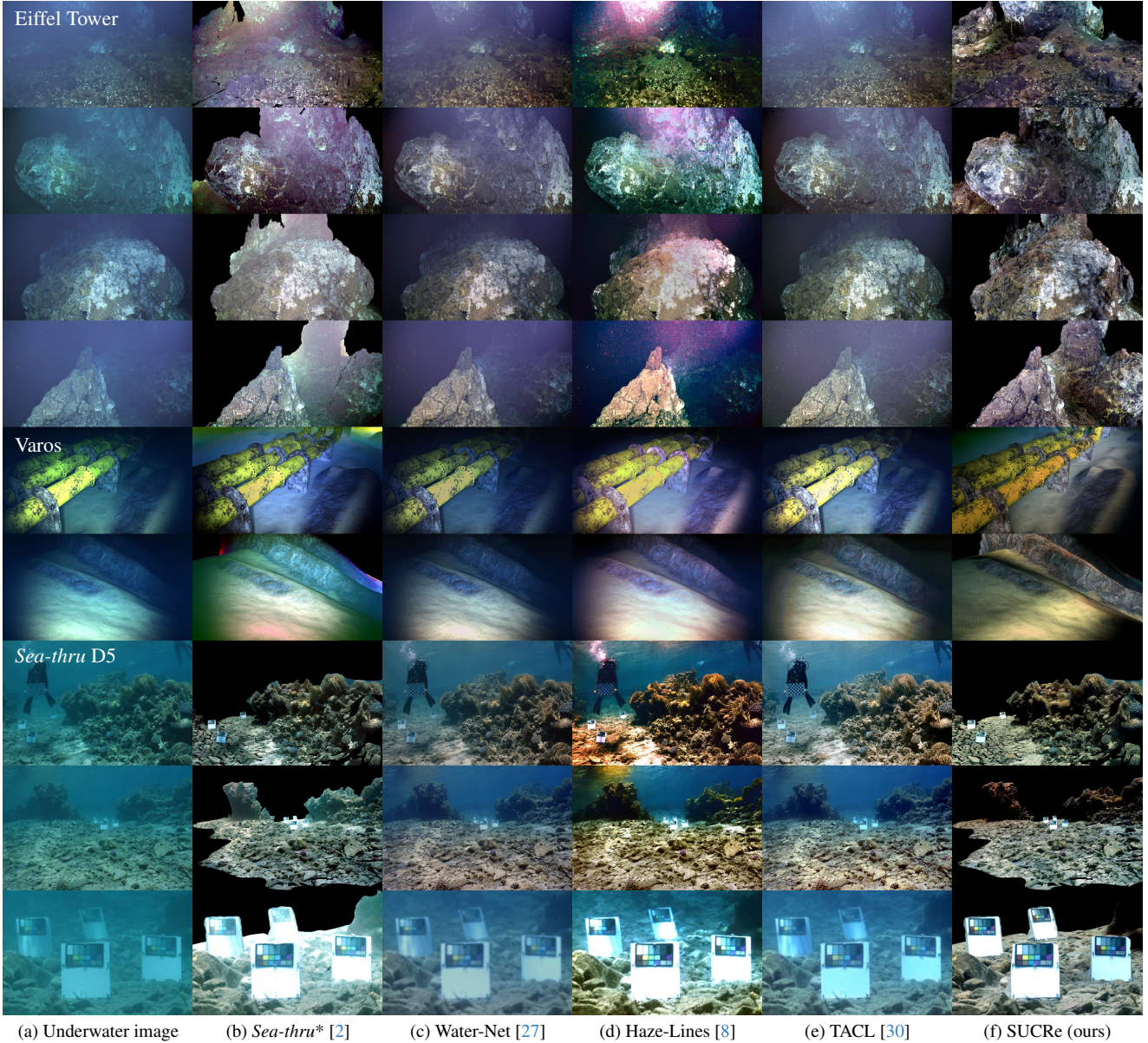
\begin{figure*}
  \centering
  \begin{subfigure}{0.16666666666\linewidth}
    \begin{picture}(60,60)
    \put(0,0){\includegraphics[width=\linewidth]{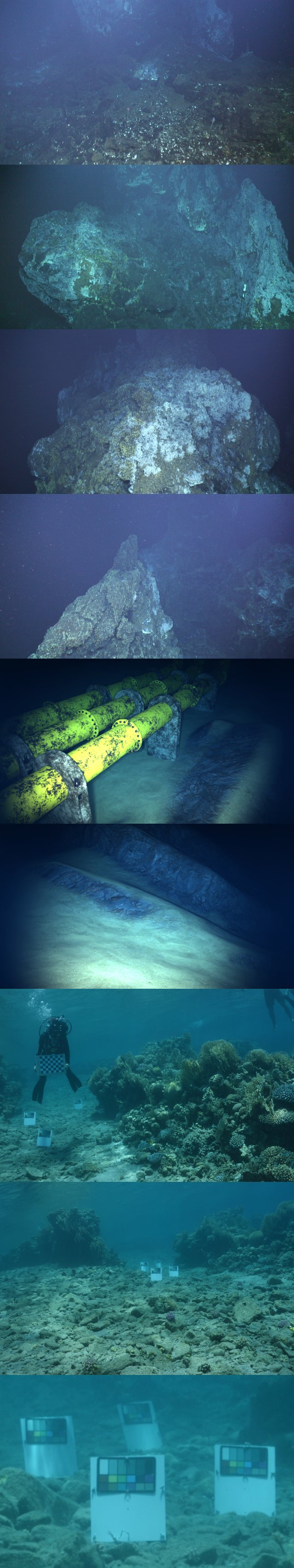}}
    \put(5,432){\color{white} \small Eiffel Tower}
    \put(5,246){\color{white} \small Varos}
    \put(5,154){\color{white} \small \textit{Sea-thru} D5}
    \end{picture}
    \caption{Underwater image}
  \end{subfigure}%
  \begin{subfigure}{0.16666666666\linewidth}
    \includegraphics[width=\linewidth]{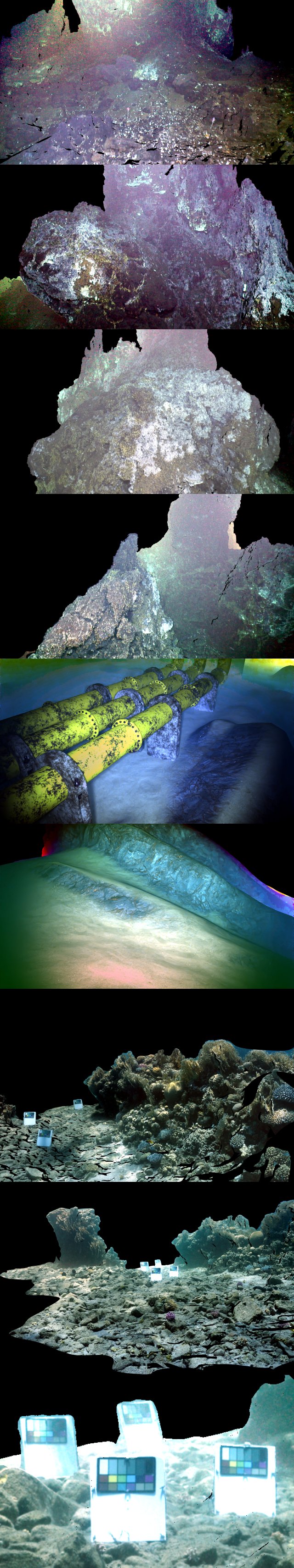}
    \caption{\textit{Sea-thru}*~\cite{akkaynak2019seathru}}
  \end{subfigure}%
  \begin{subfigure}{0.16666666666\linewidth}
    \includegraphics[width=\linewidth]{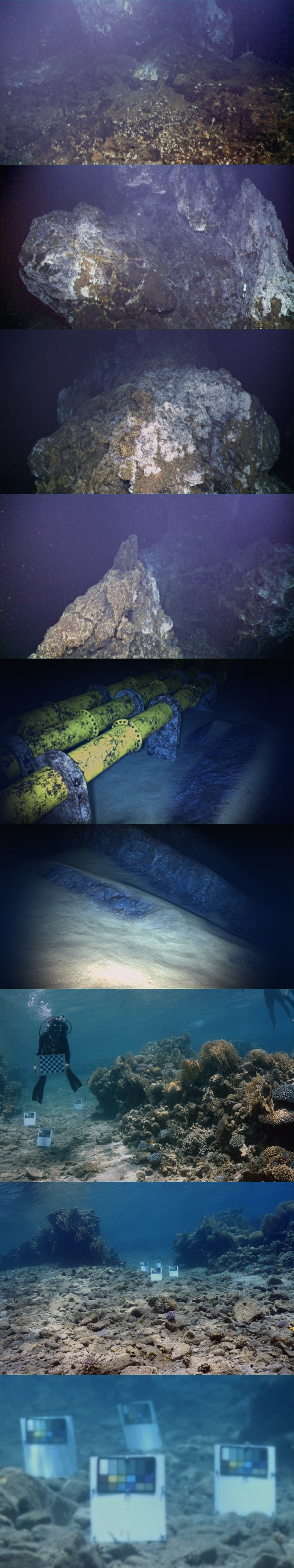}
    \caption{Water-Net~\cite{li2020benchmark}}
  \end{subfigure}%
  \begin{subfigure}{0.16666666666\linewidth}
    \includegraphics[width=\linewidth]{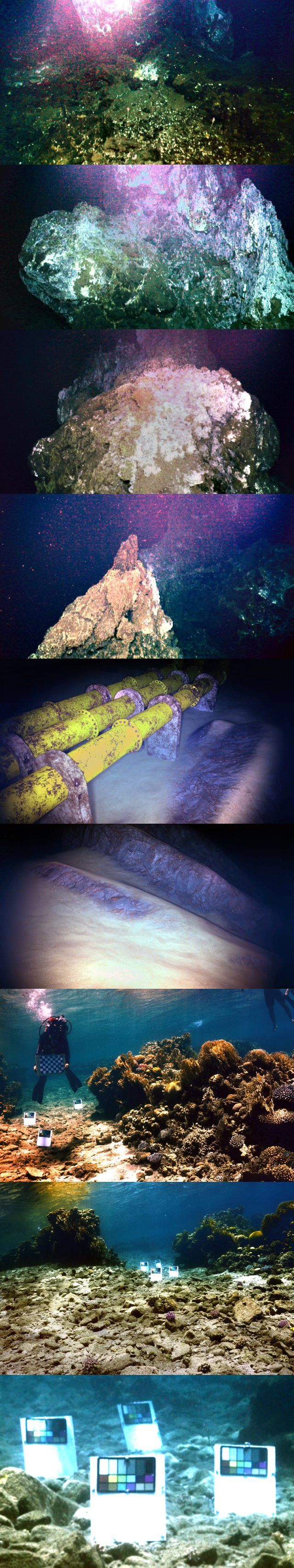}
    \caption{Haze-Lines~\cite{berman2021haze}}
  \end{subfigure}%
  \begin{subfigure}{0.16666666666\linewidth}
    \includegraphics[width=\linewidth]{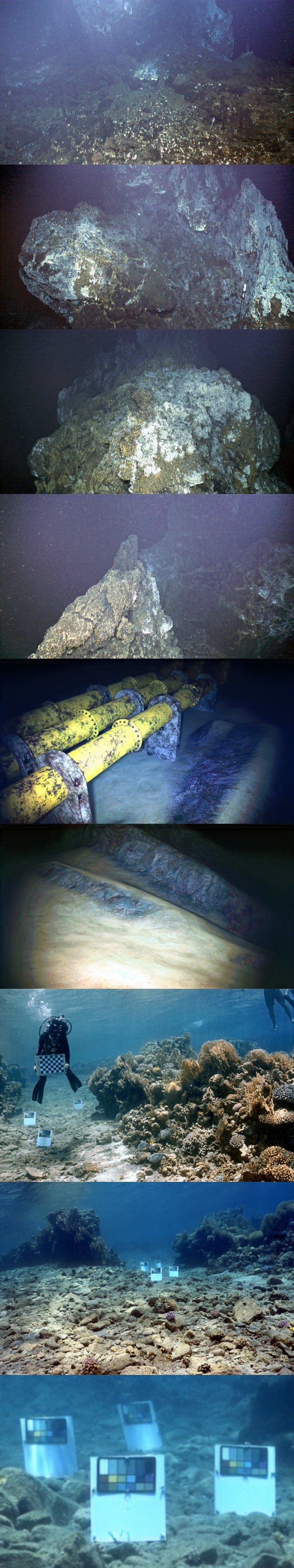}
    \caption{TACL~\cite{liu2022tacl}}
  \end{subfigure}%
  \begin{subfigure}{0.16666666666\linewidth}
    \includegraphics[width=\linewidth]{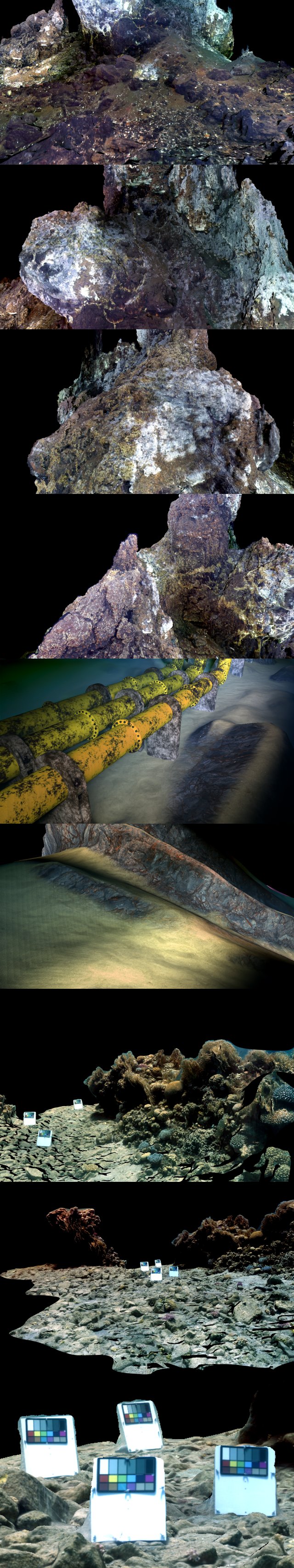}
    \caption{SUCRe (ours)}
  \end{subfigure}%
  \caption{\textbf{Visual inspection.} Restoration results of different methods in diverse underwater scenarios. The proposed approach successfully recovers colors of distant elements where other methods struggle. Zooming in on the images is encouraged. As mentioned in \cref{fig:wow}, pixels without depth information are rendered black for SUCRe and \textit{Sea-thru}*.}
  \label{fig:results}
\end{figure*}

\paragraph{Varos} is a synthetic deep-sea dataset embedding 4,715 images that were rendered with Blender within a simulated underwater setting~\cite{zwilgmeyer2021varos}. The dataset benefits from Blender's ray tracing technology, which enables the simulation of scattering and attenuation effects that are commonly observed in underwater scenes. A notable feature of Varos is its provision of reference images under uniform lighting conditions, which are useful for assessing the accuracy of color restoration techniques. These reference images offer a consistent baseline for comparison and facilitate the measurement of standard metrics such as PSNR and SSIM.
\paragraph{\textit{Sea-thru} D5} is composed of 43 raw images captured under natural light conditions, along with their corresponding distance maps obtained through SfM~\cite{akkaynak2019seathru}. The scene contains four color charts with known patterns positioned throughout the scene. These charts, visible in the last row of \cref{fig:results}, serve as ground truth for computing metrics used to evaluate the performance of color restoration algorithms.
\paragraph{Eiffel Tower} consists of a ROV dive to the Eiffel Tower hydrothermal chimney~\cite{matabos2015eiffel}. Due to the depth of the vent, which is around 1,700 meters, the ROV was equipped with an artificial lighting system since no light from the surface reaches such depths. The dataset comprises 4,875 images that were extracted from the ROV's video feed.
\paragraph{Submarine wreck} comprises 4,595 images extracted from a ROV's video feed during a dive to a submarine wreck at a depth of around 1,150 meters. Similar to the Eiffel Tower dataset, the ROV was equipped with an artificial lighting system to illuminate the wreck due to the lack of natural light at such depths.

\subsection{Implementation}
Methods such as \textit{Sea-thru}~\cite{akkaynak2019seathru} have been designed to work on raw image files. However, in most real world applications, the images available have been archived in viewable formats with low dynamic range. Despite such sub-optimal conditions to perform color restoration, our approach is able to tackle this issue. Therefore, all experiments have been run on 8-bits images, which were processed from the raw images of the \textit{Sea-thru} D5 dataset.

\begin{figure*}
  \centering
  \begin{subfigure}{0.5\linewidth}
    \includegraphics[width=\linewidth]{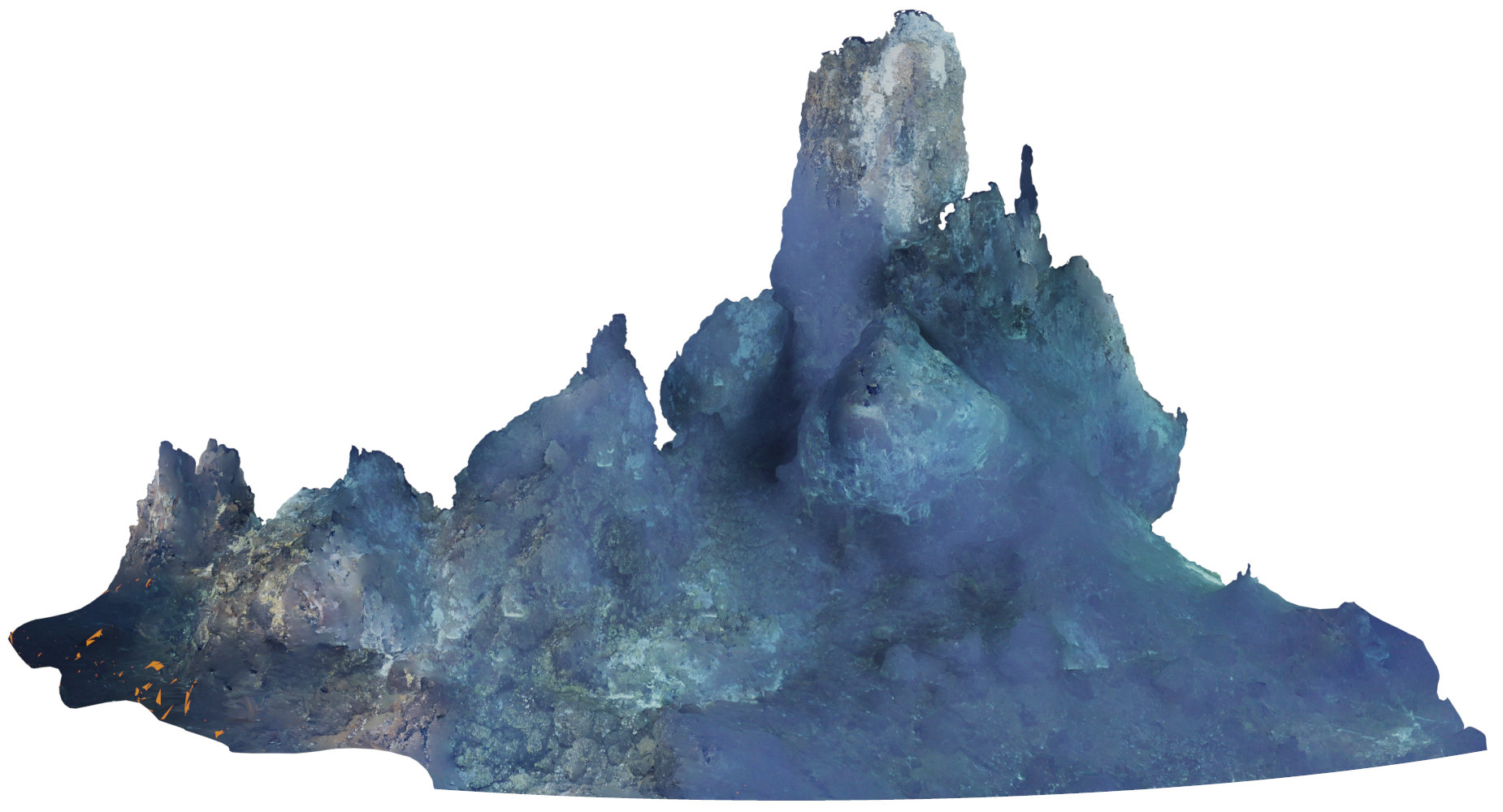}
    \caption{3D model textured with underwater images}
  \end{subfigure}%
  \begin{subfigure}{0.5\linewidth}
    \includegraphics[width=\linewidth]{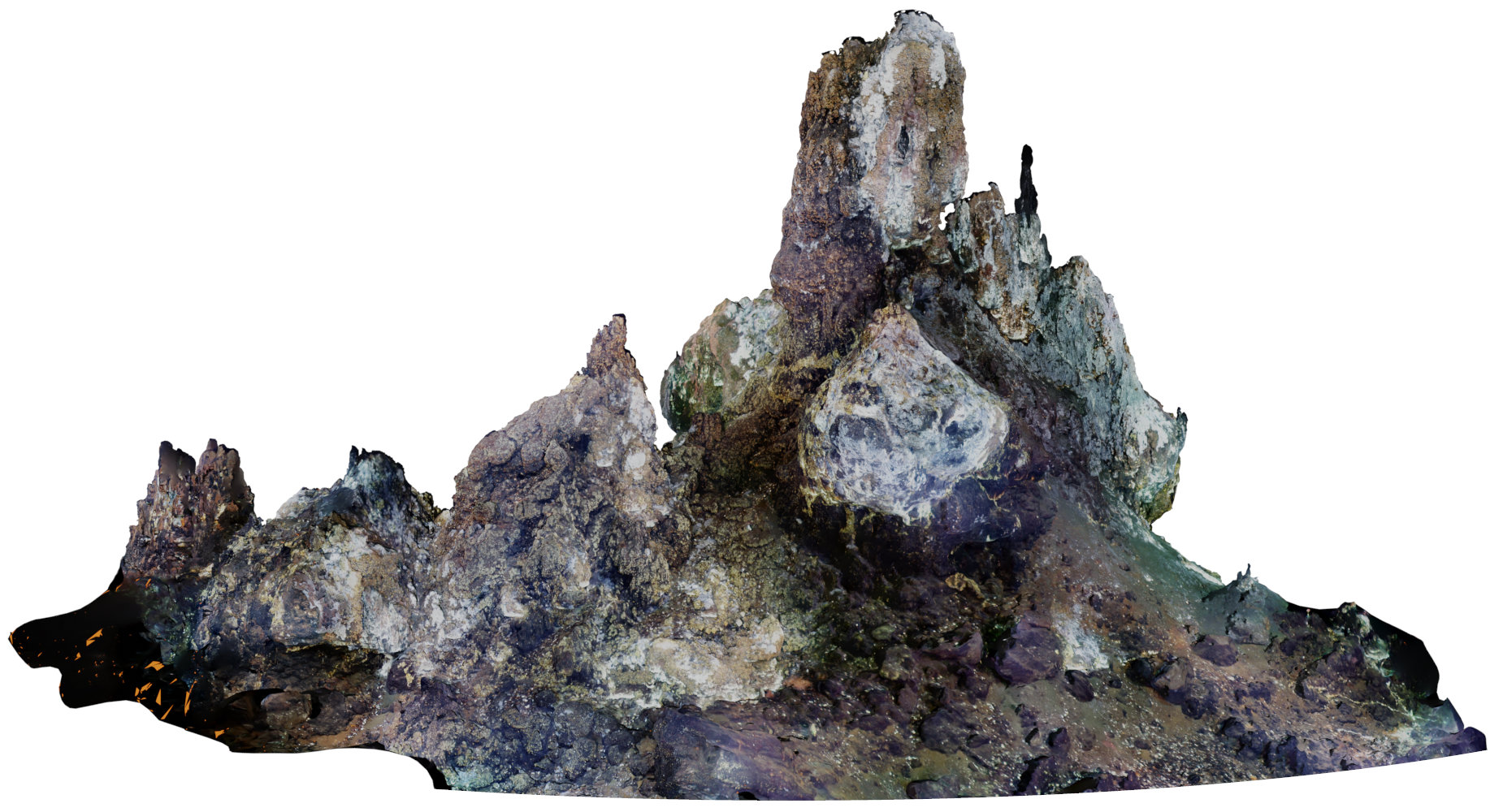}
    \caption{3D model textured with sweetened SUCRe images}
  \end{subfigure}%
  \caption{\textbf{Texturing} the Eiffel Tower hydrothermal vent 3D model with images restored using our method results in a final model with improved visual quality, including finer details and more accurate colors compared to the original model.}
  \label{fig:texturing}
\end{figure*}

\begin{figure}[t]
    \centering
    \includegraphics[width=\linewidth]{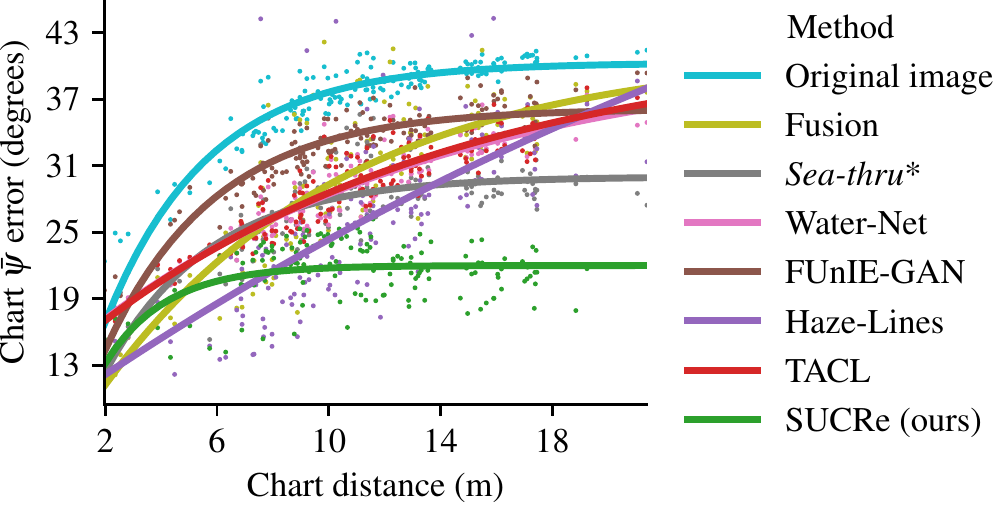}
    \caption{\textbf{Chart $\pmb{\bar{\psi}}$ error vs. distance.} Compared to other methods, SUCRe demonstrates lower and more consistent $\bar{\psi}$ errors independently of the distance of the color chart.}
    \label{fig:psidist}
\end{figure}

\paragraph{\textit{Sea-thru}*:} Due to the lack of a publicly available implementation, the \textit{Sea-thru} algorithm~\cite{akkaynak2019seathru} was re-implemented to enable its comparison with our proposed method. We adapted the implementation to work on 8-bits images, and this revised version is referred to as \textit{Sea-thru}*.

\paragraph{Initialization:} Solving the system described by \cref{eq:sfmseathru_likelihood} requires only a coarse initialization. In our experiments the following naive initial solution is sufficient: $\pmb{J} = \pmb{I}$ and $\beta = B = \gamma = 0.1$. Nevertheless, we may also use the parameters estimated with \textit{Sea-thru}* or other methods.

\begin{figure}[t]
    \centering
    \includegraphics[width=\linewidth]{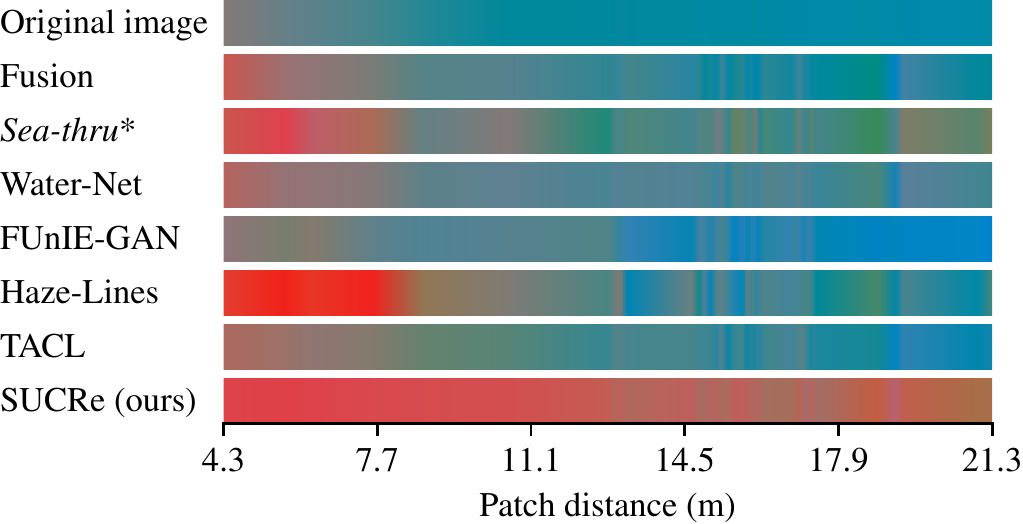}
    \caption{\textbf{Hue vs. distance.} Tracking the hue value of the red color patch at different distances on \textit{Sea-thru} D5 dataset shows that our method outperforms other approaches in maintaining stable hue values. Similar studies on other color patches are available in \Cref{ap:hue}.}
    \label{fig:red}
\end{figure}

\paragraph{Optimization:} To optimize the many parameters involved in jointly estimating the restored image along with the UIFM parameters, we use gradient descent with an Adam optimizer~\cite{kingma2015adam}. Each step of the gradient descent is computed using all matched observations by minimizing the function described by \cref{eq:sfmseathru_likelihood}. Specifically, we perform 200 optimization steps with a learning rate of 0.05.

\paragraph{Normalization:}
Most color restoration methods need to perform white balancing as a final processing step. This is usually done using the Gray World Hypothesis algorithm~\cite{buchsbaum1980grayworld}. Instead, we choose to apply a simple channel-wise histogram stretching.

\paragraph{Processing time:} The full pipeline processing time depends on the number of images in the dataset and evolves linearly with the number of observations in SUCRe's optimization. Details are provided in \Cref{ap:time}.

\subsection{Quantitative evaluation}

Evaluating the performance of underwater color restoration methods is a challenging task. Ground truth restored colors are generally unavailable for real-world underwater images. Nonetheless, synthetic datasets like Varos or images containing color charts like those included in the \textit{Sea-thru} D5 dataset can be employed to provide reference values. This allows for the use of so called full-reference metrics~\cite{li2020benchmark}. We may also employ proposed no-reference metrics to assess other factors like contrast and saturation~\cite{yang2015uciqe, panetta2016uiqm}. However, recent literature has raised doubts about the ability of the latter to accurately measure the correction of physical phenomena such as attenuation and scattering~\cite{li2020benchmark, jiang2022evaluation}.

PSNR and SSIM are full-reference measures of image similarity that are particularly useful when entire ground truth restored images are available. UCIQE~\cite{yang2015uciqe} and UIQM~\cite{panetta2016uiqm} are commonly used no-reference metrics for evaluating the visual quality of restored underwater images. The CIEDE2000 ($\Delta E_{00}$) formula was developed by the International Commission on Illumination to evaluate color differences~\cite{sharma2005ciede2000} and is commonly used in underwater color restoration~\cite{ancuti2017dehazing, li2021transmission}. We hereby compute it between the restored and expected color patches of the \textit{Sea-thru} D5 dataset. The $\bar{\psi}$ error was introduced by Berman \etal~\cite{berman2021haze} and is designed specifically for images with color charts of known colors distributed throughout the scene. For a given color chart in an image, the $\bar{\psi}$ error is defined as the average angle in RGB space between grayscale patches and a pure gray color. We redefine the error to take into account all twelve color patches in the color calibration charts used in the \textit{Sea-thru} D5 dataset:
\begin{equation}
  \bar{\psi} = \frac{1}{12} \sum_{p \in P} \cos^{-1} \left( \frac{\pmb{I}_p \cdot \pmb{E}_p}{\Vert \pmb{I}_p \Vert \cdot \Vert \pmb{E}_p \Vert} \right),
  \label{eq:eval}
\end{equation}
where $P$ is a set containing pixel indices of the twelve color patches in the given color chart and $\pmb{E}_p$ denotes the expected RGB values of the color patch with pixel index $p$.

\Cref{tab:results} reports PSNR, SSIM, UCIQE~\cite{yang2015uciqe} and UIQM~\cite{panetta2016uiqm} metrics on the Varos dataset, as well as $\bar{\psi}$ error in degrees~\cite{berman2021haze} and CIEDE2000 ($\Delta E_{00}$) color difference~\cite{sharma2005ciede2000} on the \textit{Sea-thru} D5 dataset. Results show that the proposed approach significantly outperforms other methods on every full-reference metric. Similarly to previous works~\cite{li2020benchmark, jiang2022evaluation}, we find that there is little correlation between PSNR/SSIM objective criteria and UCIQE/UIQM metrics. Furthermore, the low standard deviations on \textit{Sea-thru} D5 errors show that SUCRe retains a high performance across color chart positions, a unique quality of our method that is highlighted in \cref{fig:psidist,fig:red}.

\subsection{Qualitative evaluation}\label{sec:qualitative}

As can be seen on \cref{fig:pipeline,fig:wow,fig:results}, our method successfully restores the colors of far away elements (see last row of \cref{fig:results}). The use of multiple overlapping views overcomes the low intensities and partial loss of color information due to 8-bits quantization. In other words, our method fully exploits the virtual dynamic range augmentation offered by multiple observations of the color charts in different images. Additionally this ensures color consistency across elements of the scene at different distances from the sensor. \Cref{fig:texturing} illustrates that, when texturing a 3D mesh, restoring underwater images using our approach leads to significant improvements, including finer details as well as plausible and coherent colors.

\subsection{Additional insights}

\begin{table}
    \begin{center}
    \small
    \begin{tabularx}{\linewidth}{YYcc}
    \toprule
    \multicolumn{2}{c}{Parameters estimation} & \multirow{2}{*}{PSNR} & \multirow{2}{*}{SSIM} \\
    \cmidrule(lr){1-2}
    Single-view & Multi-view & & \\
    \midrule
    $\pmb{J}, \beta, B, \gamma$ & --- & 10.15 & 0.39\\
    $\beta, B, \gamma$ & $\pmb{J}$ & 11.32 & 0.42\\
    --- & $\pmb{J}, \beta, B, \gamma$ & 12.13 & 0.42\\
    \bottomrule
    \end{tabularx}
    \end{center}
    \caption{\textbf{Ablation study on Varos.} We show the benefits of leveraging multi-view observations for the estimation of the UIFM parameters and the restored image.\label{tab:ablation}}
\end{table}

\paragraph{Ablation study:}
\Cref{tab:ablation} investigates the impact of using multiple views when estimating different parts of the complete model. The first row shows \textit{Sea-thru}* results, where both the UIFM parameters ($\beta$, $B$ and $\gamma$) and the restored image $\pmb{J}$ are estimated using a single image. In the second row, the UIFM parameters are fixed to those obtained with \textit{Sea-thru}*, yet the restored image is obtained using multi-view observations to minimize the same error as SUCRe (\cref{eq:sfmseathru_likelihood}). The last row shows SUCRe results, where all parameters are estimated in a multi-view setting. Results show that SSIM improves mainly by the recovery of low contrast areas when $\pmb{J}$ is estimated using multiple views. While PSNR values indicate that using multi-view observations for estimating both the UIFM parameters and the restored image is required to obtain the peak performance exhibited by our approach.

\paragraph{Degradation study:}
As the underwater vehicle explores the seabed, it captures more images and acquires additional information that was not available in previous images. \Cref{tab:degradation} shows that our approach is able to consistently make use of this information to improve restoration results, yielding higher PSNR and SSIM values when presented with more candidate matching images.

\begin{table}
    \begin{minipage}{0.37\linewidth}
    \centering
    \includegraphics[width=\linewidth]{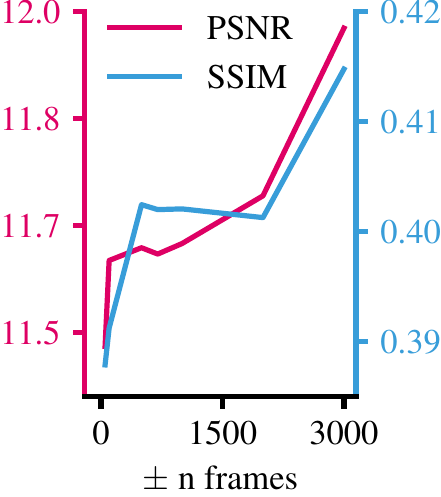}
    \end{minipage}\hfill%
    \begin{minipage}{0.63\linewidth}
    \begin{flushright}
    \small\centering
    \begin{tabular}{lcc}
    \toprule
    $\pm$ n frames &PSNR &SSIM \\
    \midrule
    50 & 11.48 & 0.39 \\
    100 & 11.61 & 0.39 \\
    500 & 11.63 & 0.40 \\
    700 & 11.62 & 0.40 \\
    1000 & 11.64 & 0.40 \\
    2000 & 11.71 & 0.40 \\
    3000 & 11.97 & 0.41 \\
    \bottomrule
    \end{tabular}
    \end{flushright}
    \end{minipage}
    \caption{\textbf{Degradation study on Varos.} We evaluate the impact of the number of sequential input frames on the performance of our approach. As the Varos dataset has a recording rate of 10fps, using $\pm$ 50 frames as input is equivalent to using 100 frames, or 10 seconds of video.\label{tab:degradation}}
\end{table}

\section{Conclusion}\label{sec:conclusion}
In this paper, we introduced SUCRe, an underwater image color restoration approach that makes use of multi-view observations to simultaneously estimate the parameters of an underwater image formation model along with the restored image. We demonstrated that the use of overlapping views allows resolving the colors of pixels that are barely visible in the target original image. It also leads to more accurate and consistent color renditions of elements at different distances from the sensor. Results on deep-sea and natural light datasets show that color restoration has much to benefit from multi-view observations of the same pixels. Furthermore, images restored with SUCRe could be used as targets to train single-view deep-learning methods. We hope this method is a step forward towards helping us view the world under the surface the same way we see the one above.
{
    \small
    \bibliographystyle{ieeenat_fullname}
    \bibliography{bibliography}
}
\clearpage
\setcounter{page}{1}
\maketitlesupplementary

\appendix

The following pages provide additional details on the experiments conducted in the main paper. \Cref{ap:ols} presents the residual analysis which justifies the use of a least squares estimator. \Cref{ap:fitting} illustrates how the underwater image formation model used in the paper fits deep-sea images. \Cref{ap:time} shows the linear trend in processing time with respect to number of observations of the reference implementation of our approach. \Cref{ap:video} describes the accompanying videos. \Cref{ap:hue} provides additional plots similar to \cref{fig:red}, tracking the hue values of all remaining color patches at different distances.

\section{Least squares}\label{ap:ols}

\begin{figure*}[!ht]
  \centering
  \begin{subfigure}{0.333333333\linewidth}
    \includegraphics[width=\linewidth]{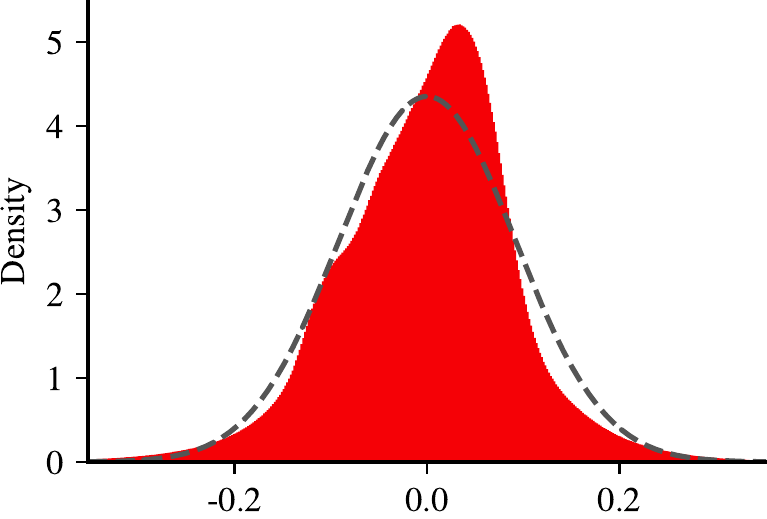}
    \caption{Residual errors on red channel}
  \end{subfigure}%
  \begin{subfigure}{0.333333333\linewidth}
    \includegraphics[width=\linewidth]{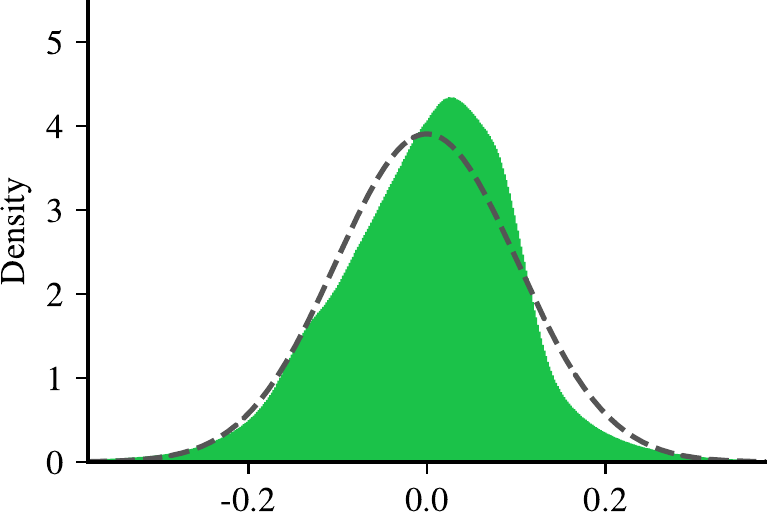}
    \caption{Residual errors on green channel}
  \end{subfigure}%
  \begin{subfigure}{0.333333333\linewidth}
    \includegraphics[width=\linewidth]{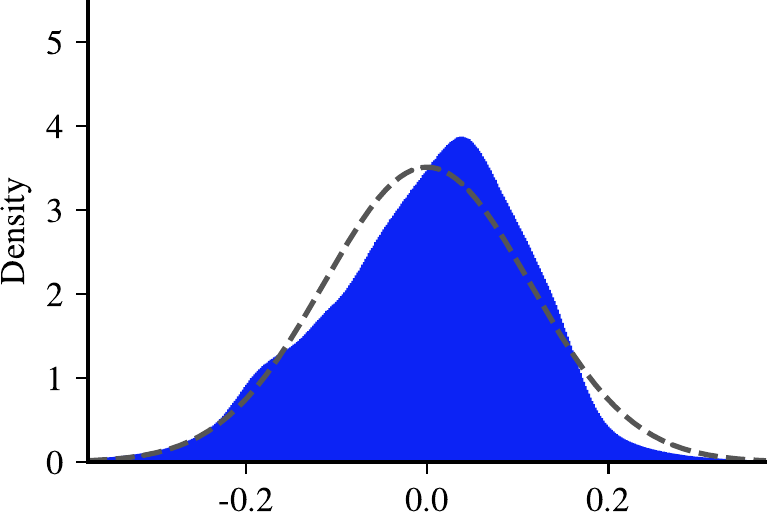}
    \caption{Residual errors on blue channel}
  \end{subfigure}%
  \caption{\textbf{Residual errors distributions.} Histograms of residual errors on a deep-sea image (\cref{fig:wow}) illustrate that their distribution is close to a Normal.}
  \label{fig:residuals}
\end{figure*}

\begin{figure*}[!ht]
  \centering
  \begin{subfigure}{0.333333333\linewidth}
    \includegraphics[width=\linewidth]{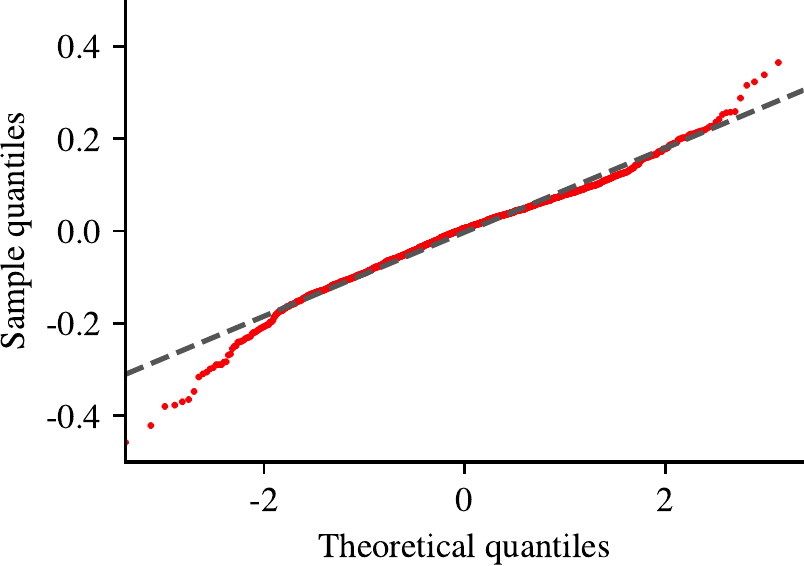}
    \caption{Normal plot on red channel}
  \end{subfigure}%
  \begin{subfigure}{0.333333333\linewidth}
    \includegraphics[width=\linewidth]{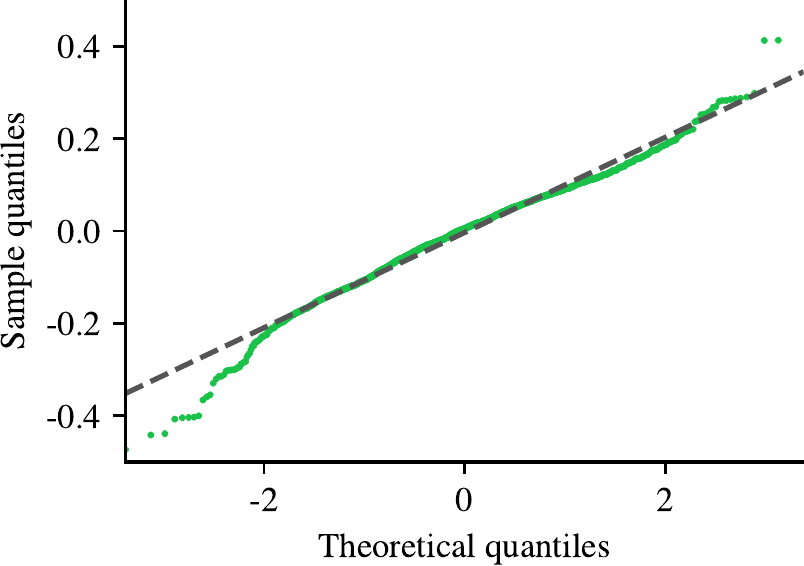}
    \caption{Normal plot on green channel}
  \end{subfigure}%
  \begin{subfigure}{0.333333333\linewidth}
    \includegraphics[width=\linewidth]{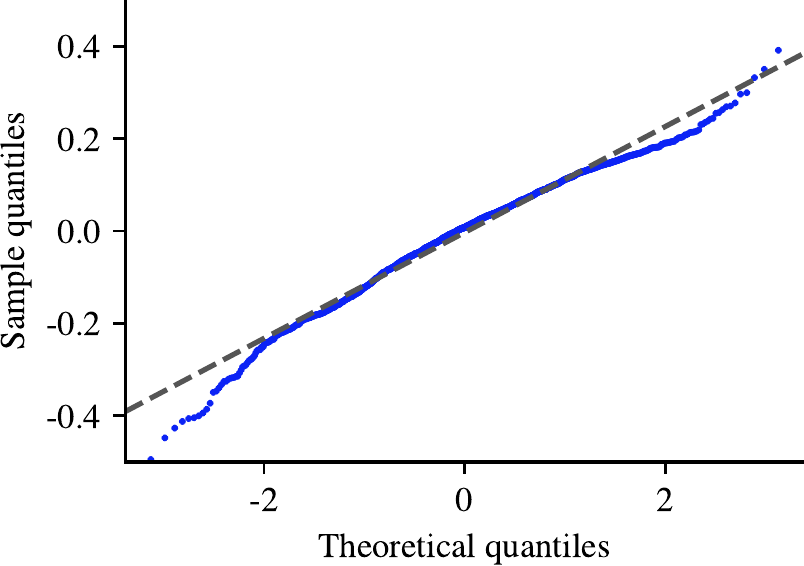}
    \caption{Normal plot on blue channel}
  \end{subfigure}%
  \caption{\textbf{Normal probability plots.} The x-axis refers to the expected quantiles of a normal distribution, while the y-axis corresponds to the actual quantiles of the residuals. Points following a straight line suggest that the data follows a Normal distribution. Plots on all three channels indicate that the residuals are close to being normally distributed.}
  \label{fig:probplot}
\end{figure*}

\begin{figure*}[!ht]
  \centering
  \begin{subfigure}{0.333333333\linewidth}
    \includegraphics[width=\linewidth]{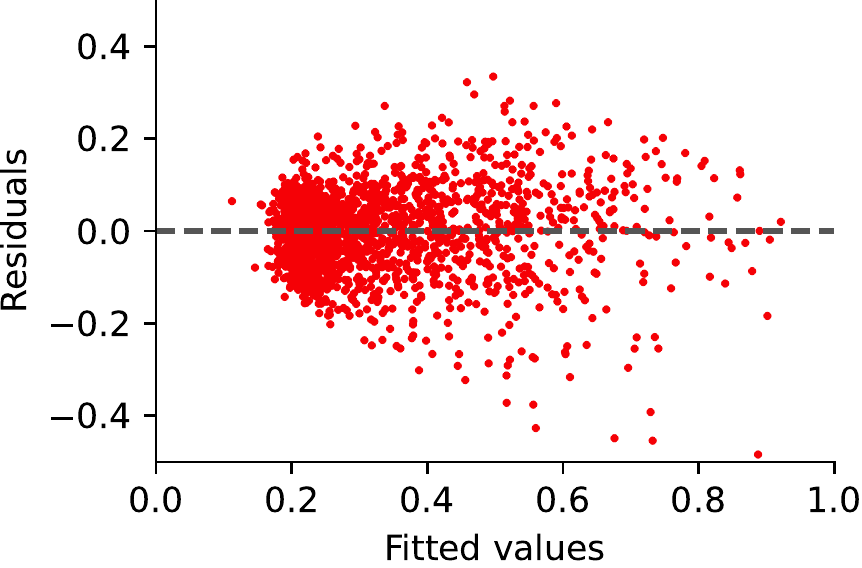}
    \caption{Residuals vs. fits on red channel}
  \end{subfigure}%
  \begin{subfigure}{0.333333333\linewidth}
    \includegraphics[width=\linewidth]{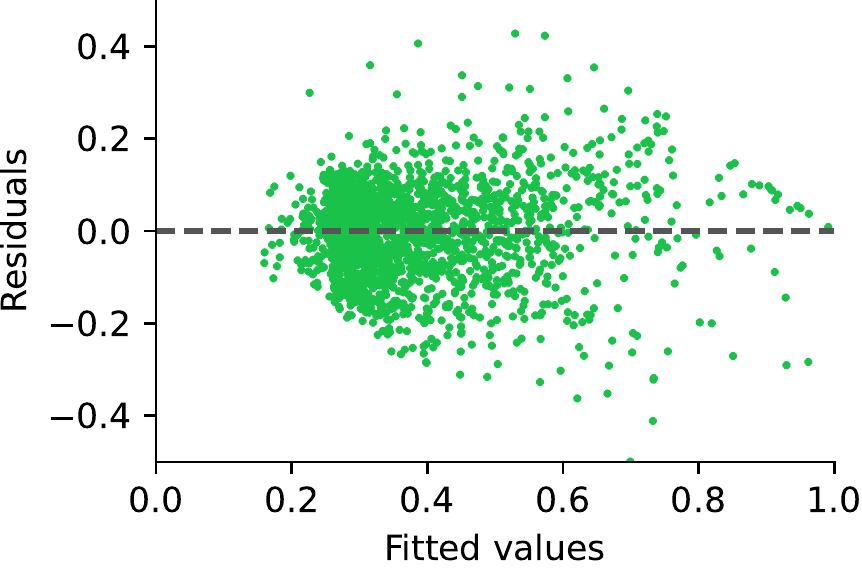}
    \caption{Residuals vs. fits on green channel}
  \end{subfigure}%
  \begin{subfigure}{0.333333333\linewidth}
    \includegraphics[width=\linewidth]{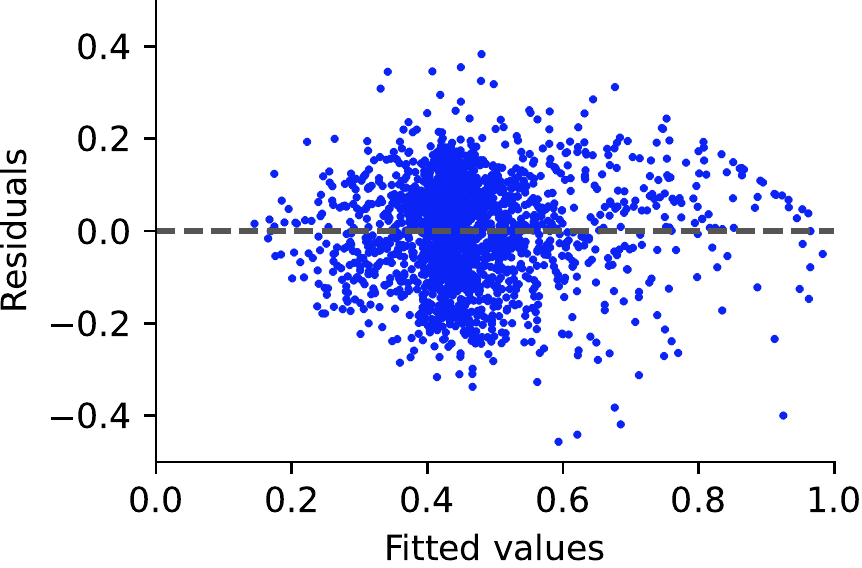}
    \caption{Residuals vs. fits on blue channel}
  \end{subfigure}%
  \caption{\textbf{Residuals vs. fits.} These plots are used to identify patterns in the residuals that may indicate problems with the model, such as non-linearity or heteroscedasticity. A well-fitting model should have residuals randomly scattered around zero independently of the fitted value, which is the case on all three channels.}
  \label{fig:fits}
\end{figure*}

Assuming that the errors between our model and measured individual per-channel pixel intensities follow a zero-centered Normal distribution, the least squares estimator leads to the maximum likelihood solution. This is the technique used in SUCRe to obtain parameters of \cref{eq:sucre_uifm}. \Cref{fig:residuals,fig:probplot,fig:fits} serve to analyze the fitting residuals on a real-world deep-sea image (\cref{fig:wow}). The deviation from normality of the residuals remains small, which supports the choice of model and estimator.

\section{Application to deep-sea images}\label{ap:fitting}
\begin{figure}[b]
    \centering
    \includegraphics[width=\linewidth]{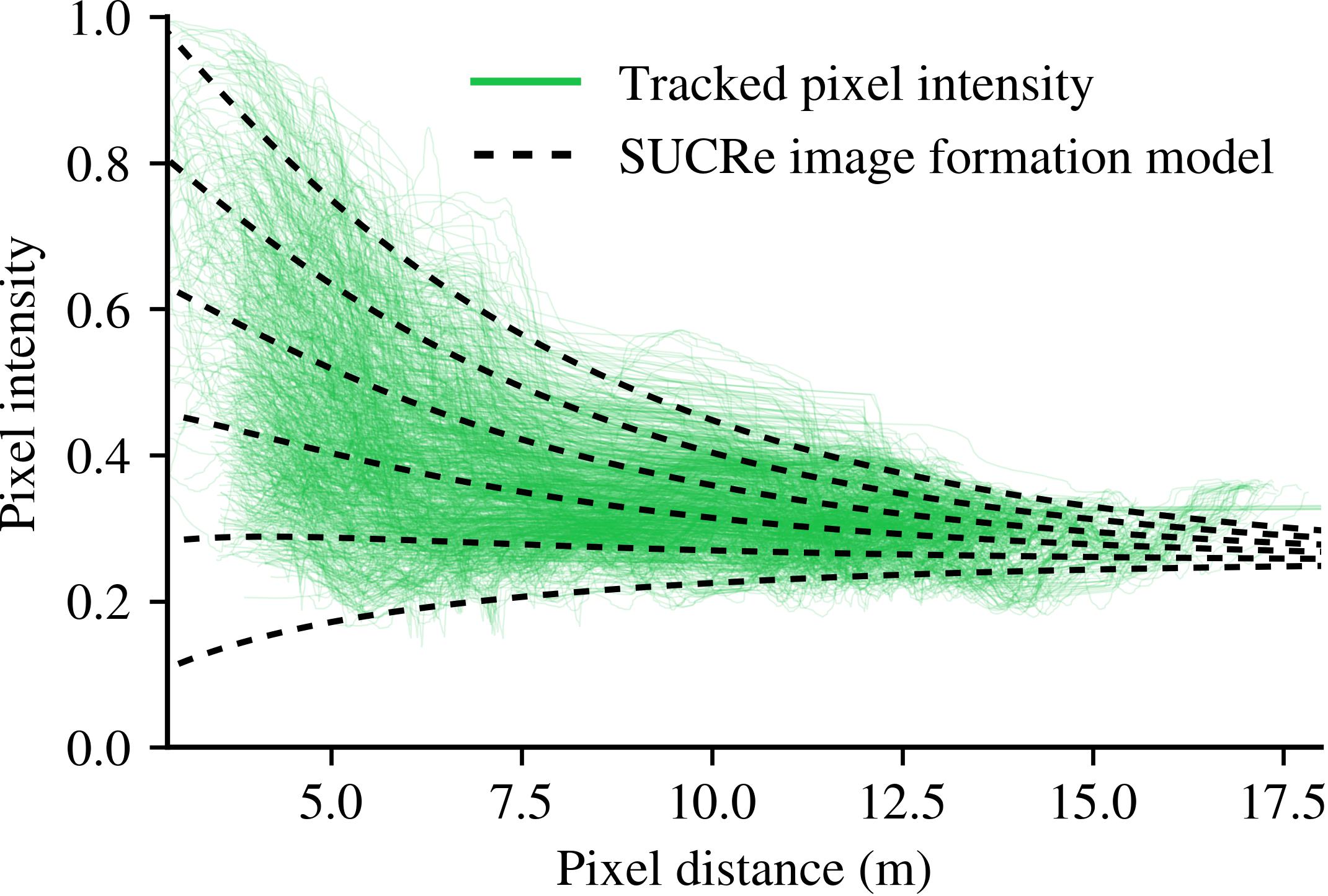}
    \caption{\textbf{Fitting the underwater image formation model on a deep-sea image.} Each green curve represents one point observed in multiple images at different distances (curves have been smoothed for visualization purposes). To illustrate how the underwater image formation model used in SUCRe fits these intensities, the black dotted lines show how different initial pixel intensities evolve with distance according to the estimated model.}
    \label{fig:fitting}
\end{figure}

\Cref{fig:fitting} shows how the underwater image formation model described by \cref{eq:sucre_uifm} fits a deep-sea image (\cref{fig:wow}). The plot illustrates that the estimated model accurately captures the evolution of observed pixel intensities with distance. This suggests that the main components described by the model, \ie, backscattering and color attenuation, adequately capture the main phenomena affecting pixel intensity changes in our specific application.

\section{Processing time}\label{ap:time}
\begin{figure}[b]
    \centering
    \includegraphics[width=\linewidth]{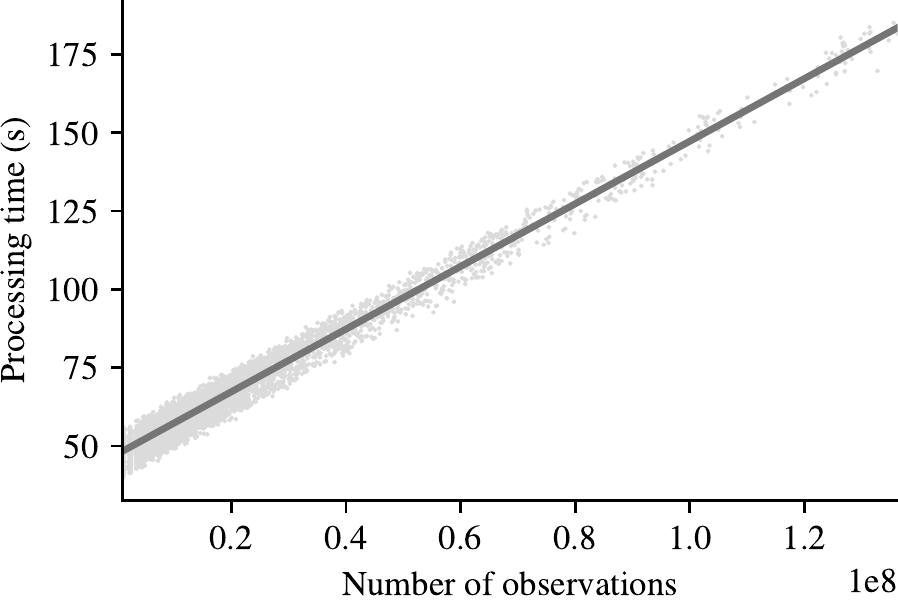}
    \caption{\textbf{Processing time vs. number of observations.} Scatter plot illustrating the relation between the number of observations in an image and the processing time of restoring it. Each point corresponds to an image of the Eiffel Tower dataset. The number of observations is the total number of pixels that have been paired to another pixel in the processed image. The processing time of the optimization procedure follows a very consistent linear increase with the number of observations. The intercept of approximately 50 seconds is due to the pairing step that computes pixel pairs for all 4,875 images in the dataset.}
    \label{fig:time}
\end{figure}

The processing time of our approach can be divided into two main components: \textit{i)} pairing pixels between the image to be restored and every other images; \textit{ii)} the optimization procedure described by \cref{eq:sfmseathru_likelihood}. The pairing step depends on the size of the images and the number of candidate matching images in the dataset. \Cref{fig:time} illustrates that the optimization step evolves linearly with the number of matched observations. To restore an image from a dataset comprising 4,875 images with a resolution of 1920 by 1080 pixels, our approach takes about 50 seconds to compute pixel pairs and 1 minute and 40 seconds for the optimization procedure using 100 millions matched pixel observations, for a total of 2 minutes and 30 seconds processing time. All computations were performed with 32 CPU threads and a RTX A5000 GPU.

\section{Videos}\label{ap:video}
The accompanying videos\footnote{\url{https://www.youtube.com/playlist?list=PLe92vnufKoYL1fkExtsmEZULREK7AOLiW}} show that our approach yields consistent colors independently of the distance between the camera and the scene. Moreover, the videos show that Haze-Lines~\cite{berman2021haze} is not well-suited for underwater color restoration of video feeds because it relies on a per-frame estimation of Jerlov water types~\cite{solonenko2015jerlov}.  Consecutive changes in the water category selection leads to flickering images in the video. While Water-Net performs a consistent restoration across frames in the video, backscattering and color attenuation remain highly visible and low contrast areas cannot be resolved without exploiting multiple views of the scene.

\begin{figure*}[t]
  \centering
  \begin{subfigure}{0.32\linewidth}
    \includegraphics[width=\linewidth]{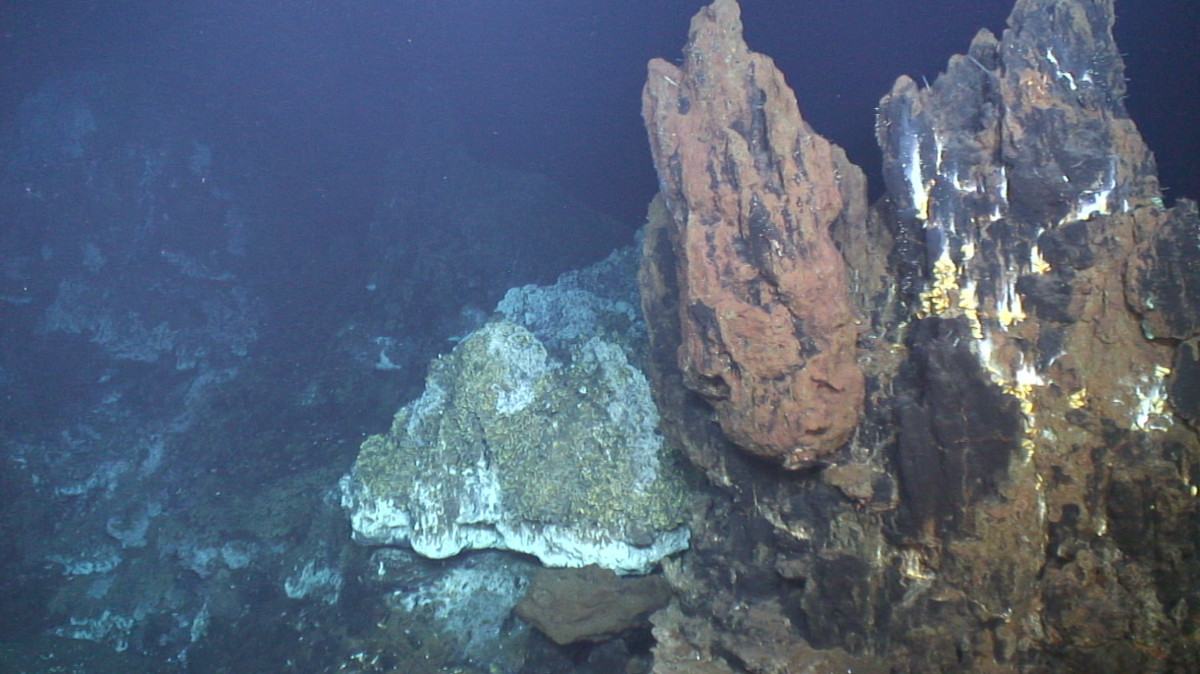}
  \end{subfigure}
  \hfill
  \begin{subfigure}{0.32\linewidth}
    \includegraphics[width=\linewidth]{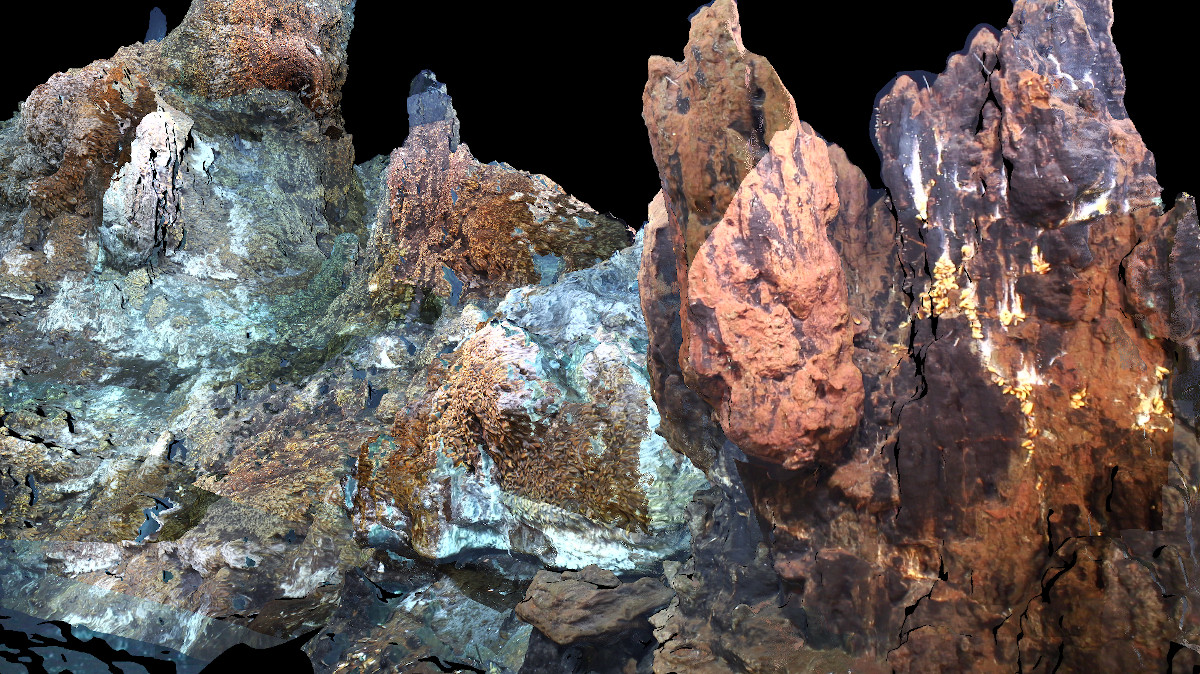}
  \end{subfigure}
  \hfill
  \begin{subfigure}{0.32\linewidth}
    \includegraphics[width=\linewidth]{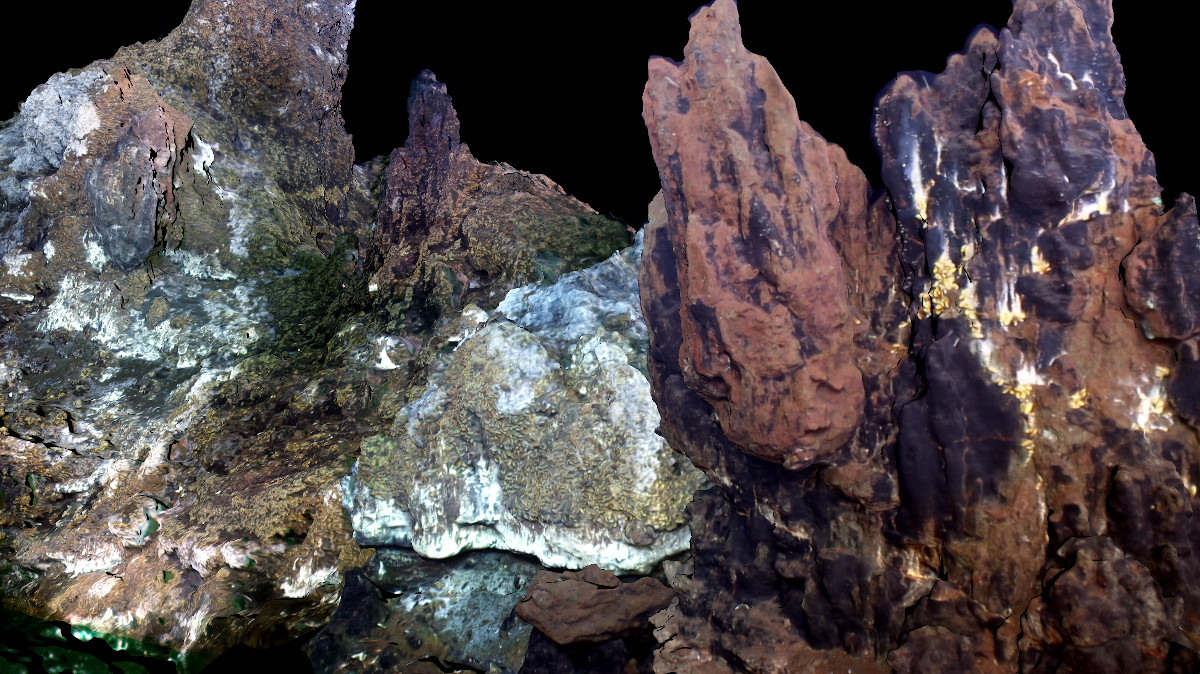}
  \end{subfigure}\\
  \vspace{8pt}
  \begin{subfigure}{0.32\linewidth}
    \includegraphics[width=\linewidth]{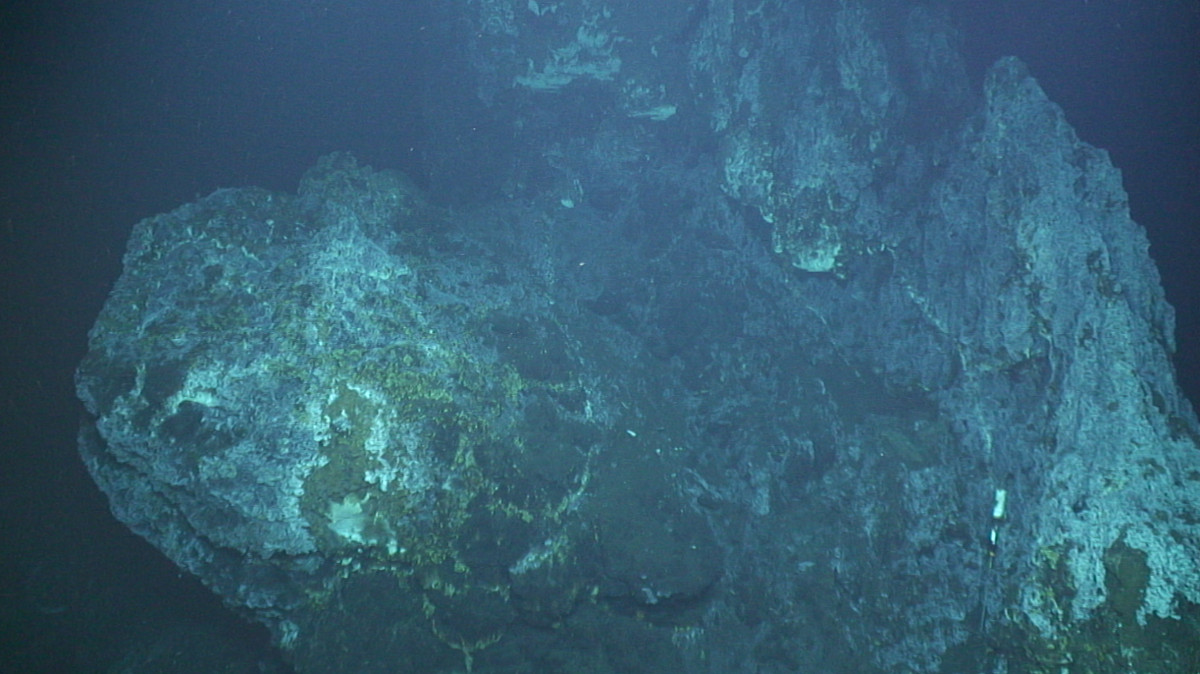}
    \caption{Original image}
  \end{subfigure}
  \hfill
  \begin{subfigure}{0.32\linewidth}
    \includegraphics[width=\linewidth]{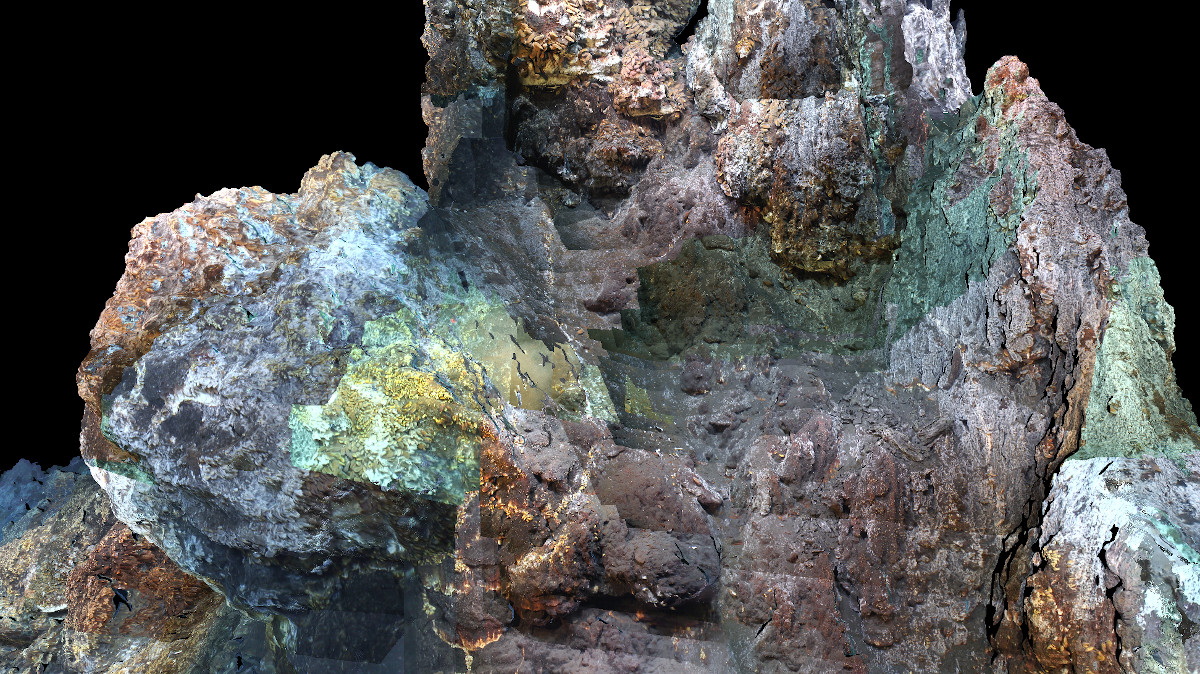}
    \caption{Stitched image}
  \end{subfigure}
  \hfill
  \begin{subfigure}{0.32\linewidth}
    \includegraphics[width=\linewidth]{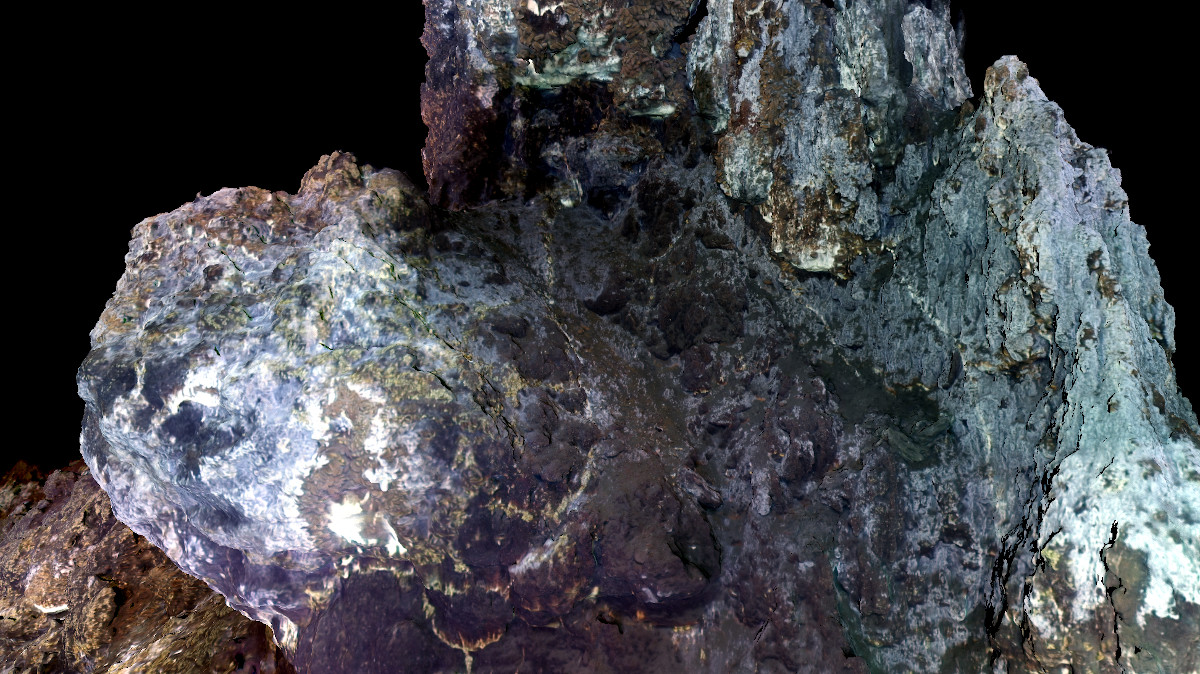}
    \caption{SUCRe}
  \end{subfigure}
  \caption{\textbf{SUCRe vs. image stitching.} Creating new images by stitching closest observations of the scene results in distorted colors and apparent seams. In contrast, SUCRe yields consistent colors across the entire image.}
  \label{fig:stitch}
\end{figure*}

\section{Limitations}
The complete process of light propagation under water is far more complex than the model presented in this paper. We make the assumption that backscatter and absorption properties are spatially and temporally consistent within all the images. In practice, the water medium experiences spatial and temporal changes, such as smoke coming out of hydrothermal chimneys, that invalidate this assumption. A complete model encompassing all light propagation phenomena that impact underwater images would require far more than the three parameters per channel used in this study.

It is also important to acknowledge that SUCRe is limited by its requirement of 3D scene information. The method is entirely dependant on the SfM quality. Poor 3D reconstruction will result in incorrect pixels pairing and impact the multi-view optimization procedure. Moreover, non-static scene elements are discarded during the SfM process and averaged out during SUCRe optimization. Because of this, SUCRe cannot render dynamic objects.

\section{SUCRe vs. image stitching}
Our approach performs a multi-view optimization using observations of the same pixels at difference distances. One could argue that SUCRe performance only stems from the availability of closer observations of the scene. To this end, we compare our method with a more naive approach: we build restored images by stitching together the intensities of each pixel's closest observation. \Cref{fig:stitch} show that this naive approach leads to distorted colors across the restored image, mainly due to viewpoint changes. Moreover, there are very apparent seams at places where different viewpoints where used to build the image. To put it simply, some parts of the stitched image result from one-meter observations, while other parts results from ten-meters observations. In contrast, the optimization procedure conducted by SUCRe smooths pixel intensities based on \cref{eq:sucre_uifm} model, including observation distance variation, and ensuring consistent colors across the entire image.

\section{Parameters analysis}
\begin{figure}[t]
  \centering
  \includegraphics[width=\linewidth]{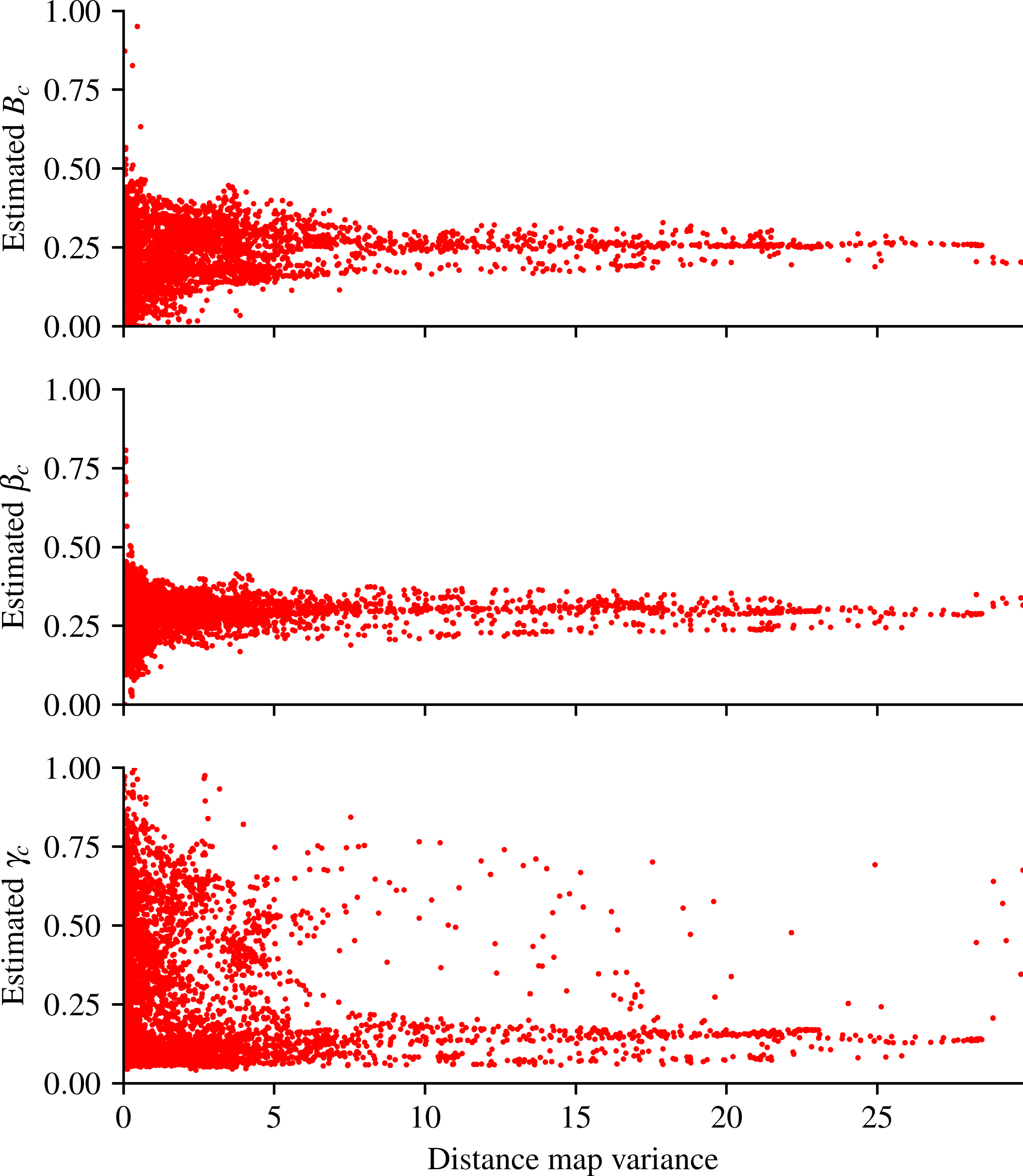}
  \caption{\textbf{Distance map variance vs. parameters estimation.} For each image of the Eiffel Tower dataset, we plot its estimated $B$, $\beta$ and $\gamma$ parameters on the red channel against the variance of its distance map.}
  \label{fig:params}
\end{figure}
The $B$, $\beta$ and $\gamma$ parameters outlined in \cref{eq:sucre_uifm} carry physical meanings. Under consistent water conditions, they are expected to exhibit similar values. In \cref{fig:params}, we present the estimation of these parameters using real-world images from the Eiffel Tower dataset. Our findings reveal that while these parameters undergo substantial changes within similar conditions, these changes are primarily correlated with the variance in the distance maps of the restored images. The parameters exhibit notable variations when the distance maps have low variance, yet they appear almost identical in cases where the distance map exhibits high variance. This behavior stems from the minimization problem articulated in \cref{eq:sfmseathru_likelihood}: in instances of very low distance variance, the available range for estimating the parameters of the model is constrained. However, this does not imply that the estimated parameters fail to capture the modeled phenomena.

\section{Tracking hue values}\label{ap:hue}

\begin{figure}[t]
    \centering
    \includegraphics[width=\linewidth]{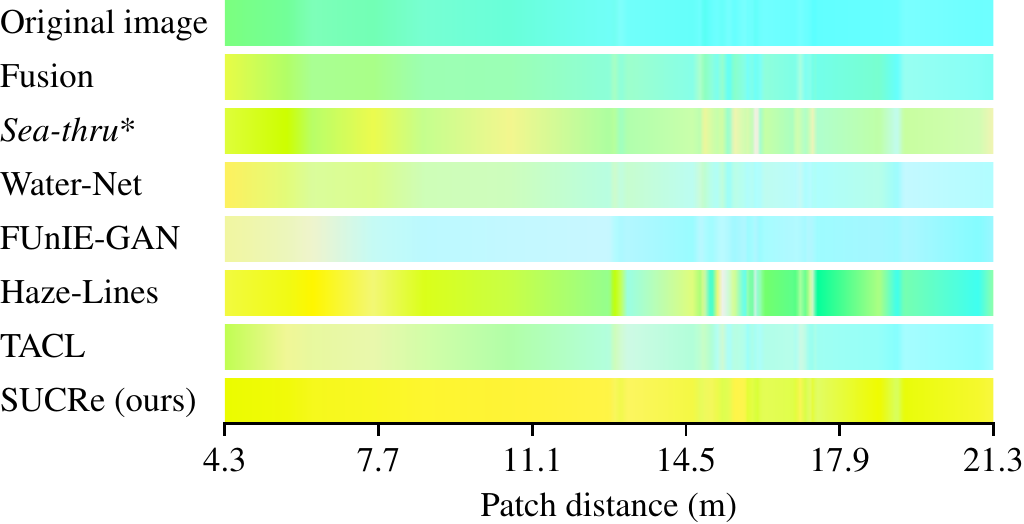}
    \caption{\textbf{Hue vs. distance} on the yellow color patch.}
    \label{fig:yellow}
\end{figure}

\begin{figure}[t]
    \centering
    \includegraphics[width=\linewidth]{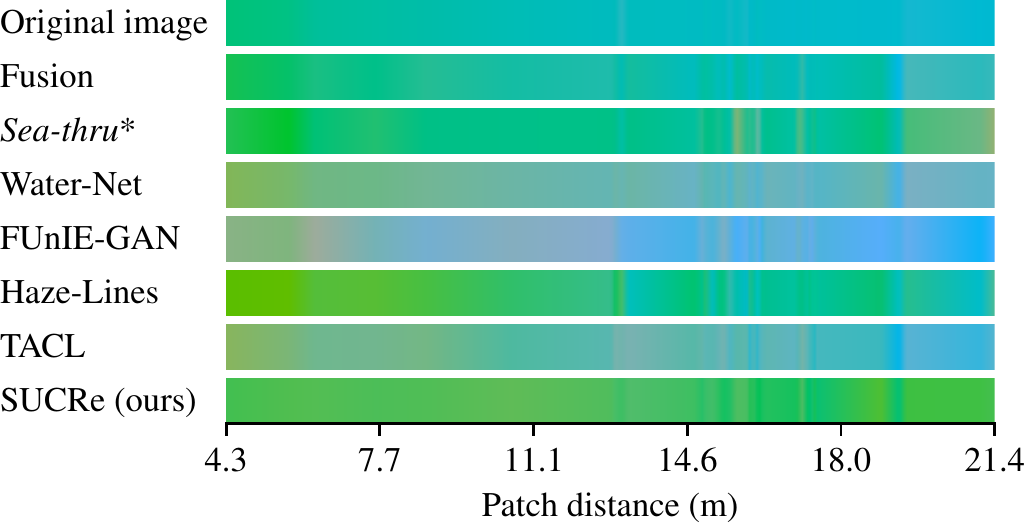}
    \caption{\textbf{Hue vs. distance} on the green color patch.}
    \label{fig:green}
\end{figure}

\begin{figure}[t]
    \centering
    \includegraphics[width=\linewidth]{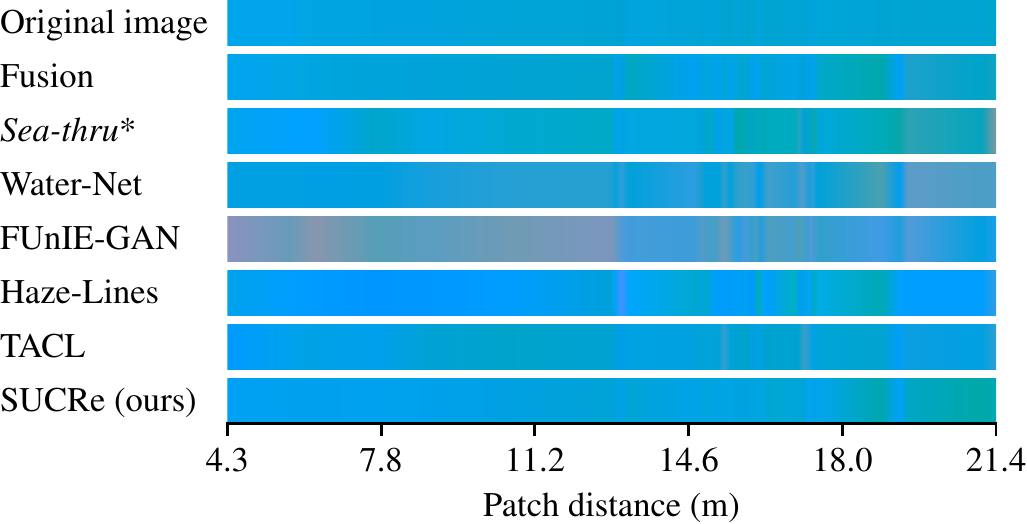}
    \caption{\textbf{Hue vs. distance} on the light blue color patch.}
    \label{fig:ligtblue}
\end{figure}

\begin{figure}[t]
    \centering
    \includegraphics[width=\linewidth]{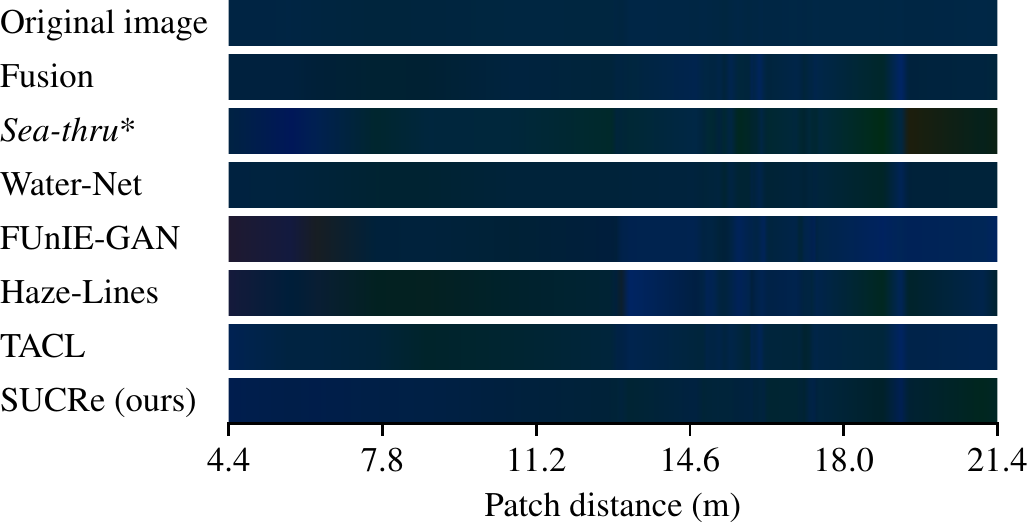}
    \caption{\textbf{Hue vs. distance} on the dark blue color patch.}
    \label{fig:darkblue}
\end{figure}

\begin{figure}[t]
    \centering
    \includegraphics[width=\linewidth]{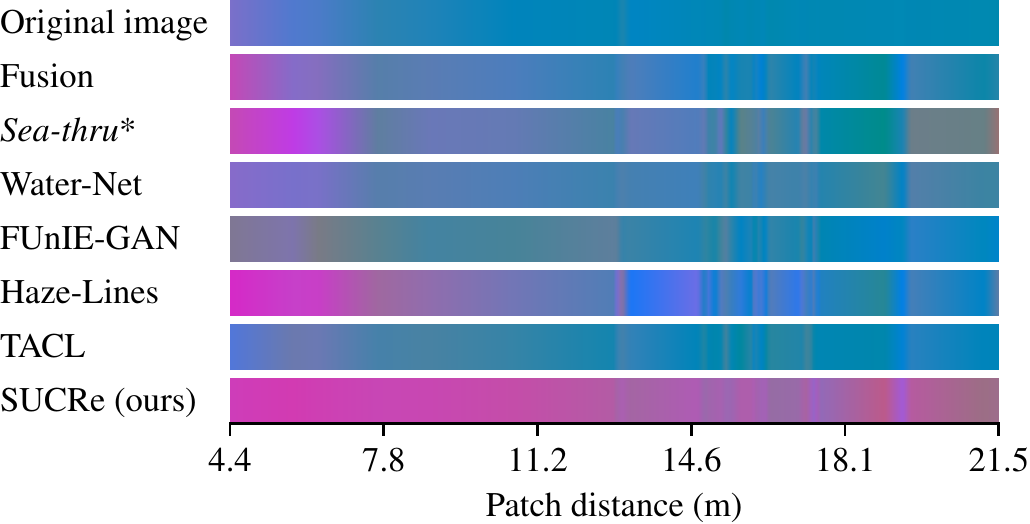}
    \caption{\textbf{Hue vs. distance} on the magenta color patch.}
    \label{fig:magenta}
\end{figure}

\begin{figure}[t]
    \centering
    \includegraphics[width=\linewidth]{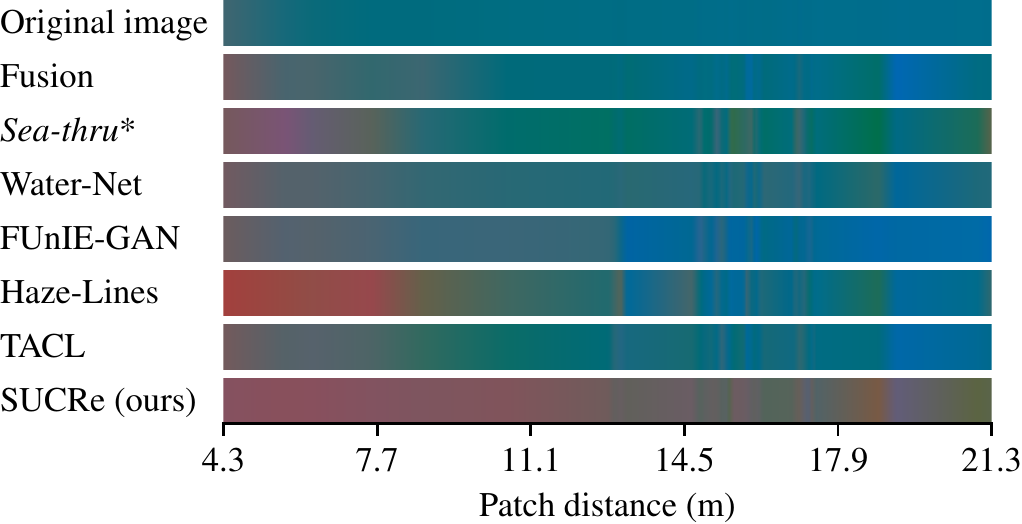}
    \caption{\textbf{Hue vs. distance} on the brick color patch.}
    \label{fig:brick}
\end{figure}

\begin{figure}[t]
    \centering
    \includegraphics[width=\linewidth]{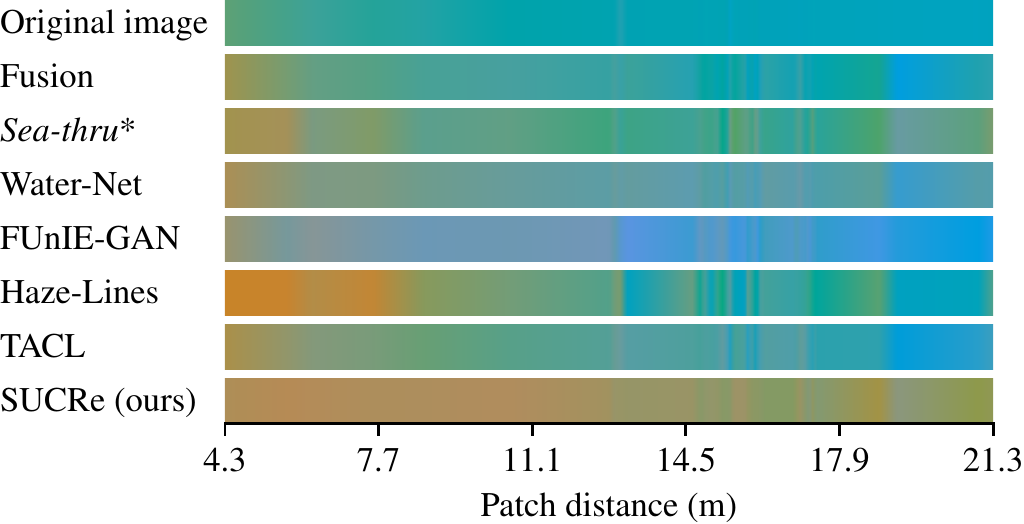}
    \caption{\textbf{Hue vs. distance} on the orange color patch.}
    \label{fig:orange}
\end{figure}

\begin{figure}[t]
    \centering
    \includegraphics[width=\linewidth]{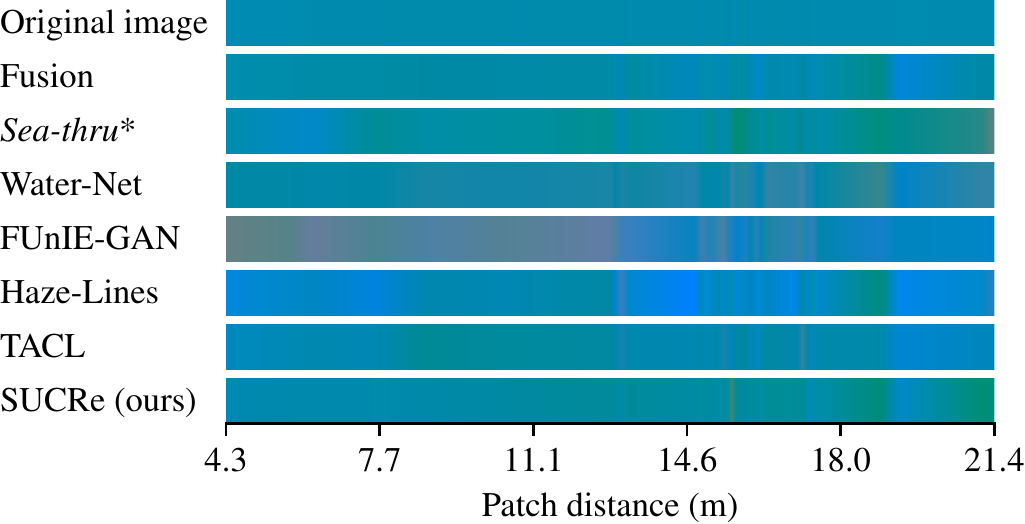}
    \caption{\textbf{Hue vs. distance} on the turchese color patch.}
    \label{fig:turchese}
\end{figure}

\begin{figure}[t]
    \centering
    \includegraphics[width=\linewidth]{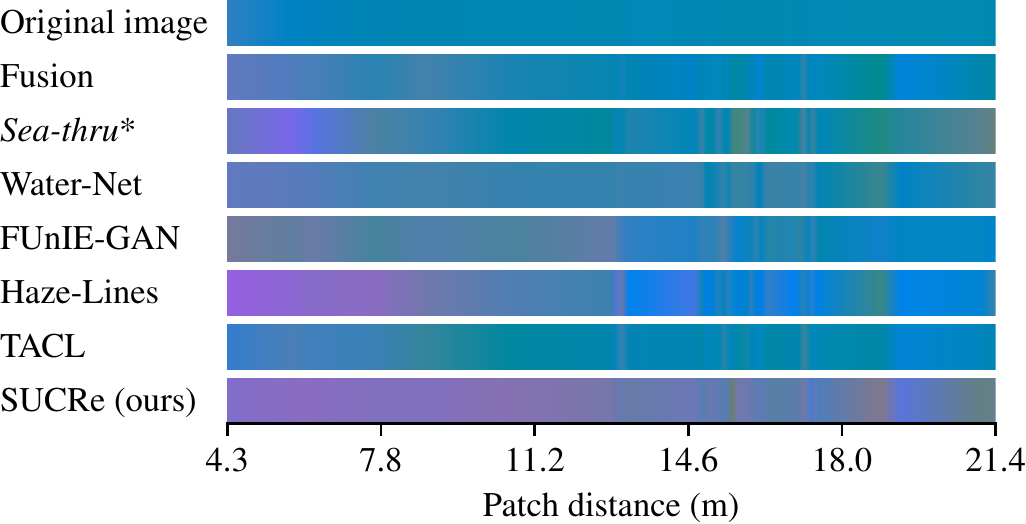}
    \caption{\textbf{Hue vs. distance} on the purple color patch.}
    \label{fig:purple}
\end{figure}

\begin{figure}[t]
    \centering
    \includegraphics[width=\linewidth]{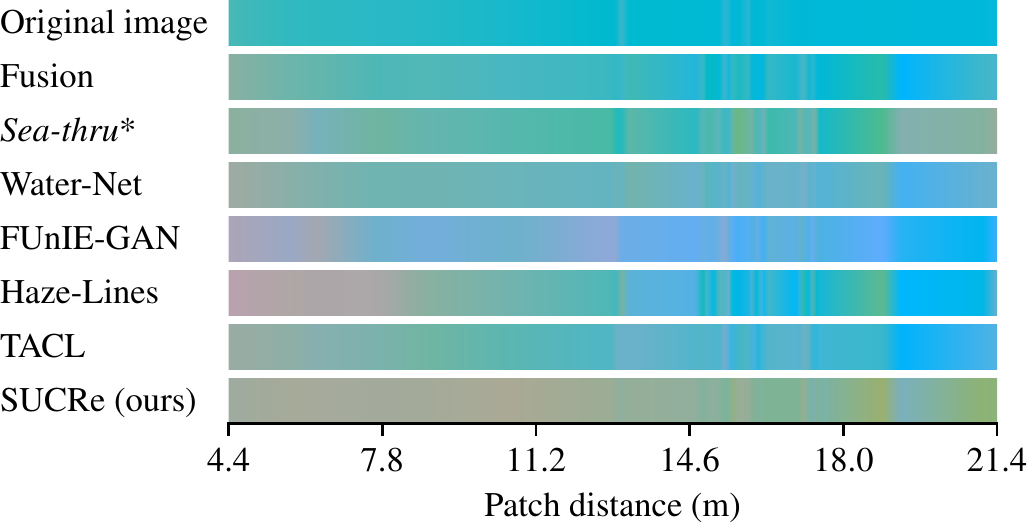}
    \caption{\textbf{Hue vs. distance} on the beige color patch.}
    \label{fig:beige}
\end{figure}

\begin{figure}[t]
    \centering
    \includegraphics[width=\linewidth]{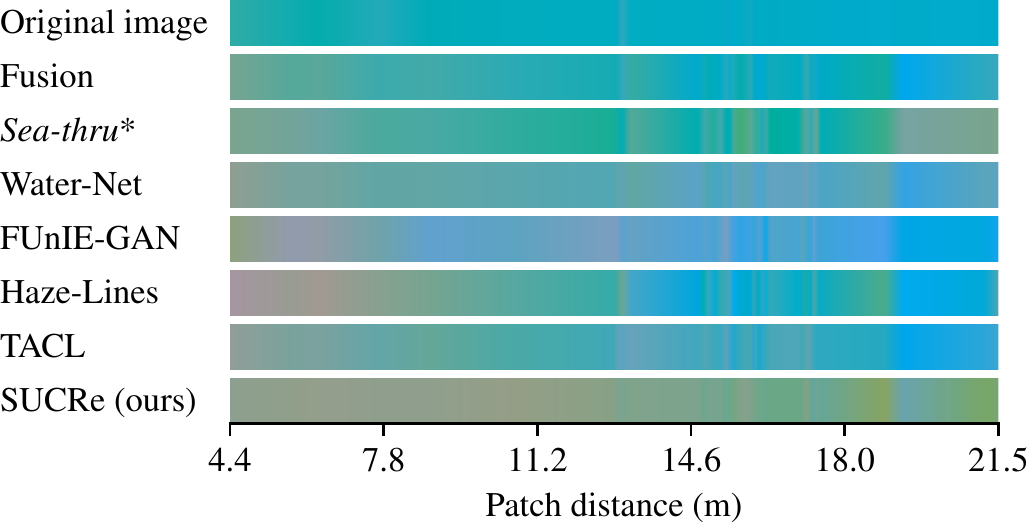}
    \caption{\textbf{Hue vs. distance} on the brown color patch.}
    \label{fig:brown}
\end{figure}

Similar to \cref{fig:red}, \cref{fig:yellow,fig:green,fig:ligtblue,fig:darkblue,fig:magenta,fig:brick,fig:orange,fig:turchese,fig:purple,fig:beige,fig:brown} show the restored hue values of all other color patches whose appearance is greatly affected under water.

\vspace{1cm}

\end{document}